\documentclass[pdflatex,sn-mathphys-num]{sn-jnl}

\usepackage{graphicx}%
\usepackage{multirow}%
\usepackage{amsmath,amssymb,amsfonts}%
\usepackage{amsthm}%
\usepackage{mathrsfs}%
\usepackage[title]{appendix}%
\usepackage{xcolor}%
\usepackage{textcomp}%
\usepackage{manyfoot}%
\usepackage{booktabs}%
\usepackage{algorithm}%
\usepackage{algorithmicx}%
\usepackage{algpseudocode}%
\usepackage{listings}%
\usepackage{lineno}
\usepackage{longtable}

\begin{document}

\title[Article Title]{ON-OFF Neuromorphic ISING Machines using Fowler-Nordheim Annealers}

\author[1]{\fnm{Zihao} \sur{Chen}}\email{czihao@wustl.edu}
\author[1]{\fnm{Zhili} \sur{Xiao}}\email{xiaozhili@wustl.edu}
\author[2]{\fnm{Mahmoud} \sur{Akl}}\email{mahmoud.akl@spinncloud.com}
\author[3]{\fnm{Johannes} \sur{Leugring}}\email{jleugering@ucsd.edu}
\author[3]{\fnm{Omowuyi} \sur{Olajide}}\email{oolajide@ucsd.edu}
\author[4]{\fnm{Adil} \sur{Malik}}\email{mam315@ic.ac.uk}
\author[5,6]{\fnm{Nik} \sur{Dennler}}\email{nd21aad@herts.ac.uk}
\author[7,8]{\fnm{Chad} \sur{Harper}}\email{chad\_harper@berkeley.edu}
\author[1]{\fnm{Subhankar} \sur{Bose}}\email{b.subhankar@wustl.edu}
\author[2,9]{\fnm{Hector A.} \sur{Gonzalez}}\email{hector.gonzalez@spinncloud.com}
\author[2]{\fnm{Mohamed} \sur{Samaali}}\email{mohamed.samaali@spinncloud.com}
\author[2]{\fnm{Gengting} \sur{Liu}}\email{gengting.liu@spinncloud.com}
\author[10]{\fnm{Jason} \sur{Eshraghian}}\email{jeshragh@ucsc.edu}
\author[11]{\fnm{Riccardo} \sur{Pignari}}\email{riccardo.pignari@polito.it}
\author[11]{\fnm{Gianvito} \sur{Urgese}}\email{gianvito.urgese@polito.it}
\author[12]{\fnm{Andreas G.} \sur{Andreou}}\email{andreou1@jhu.edu}
\author[13,14]{\fnm{Sadasivan} \sur{Shankar}}\email{Sadas.Shankar@stanford.edu}
\author[9, 15]{\fnm{Christian} \sur{Mayr}}\email{christian.mayr@tu-dresden.de}
\author[3]{\fnm{Gert} \sur{Cauwenberghs}}\email{gcauwenberghs@ucsd.edu}
\author*[1]{\fnm{Shantanu} \sur{Chakrabartty}}\email{shantanu@wustl.edu}

\affil[1]{\orgdiv{Department of Electrical and Systems Engineering}, \orgname{Washington University in St. Louis}, \orgaddress{\street{One Brookings Drive}, \city{St. Louis}, \postcode{63130}, \state{MO}, \country{USA}}}

\affil[2]{\orgname{SpiNNcloud Systems GmbH}, \orgaddress{\street{Freibergerstr. 37}, \city{Dresden}, \postcode{01067}, \country{Germany}}}

\affil[3]{\orgdiv{Department of Bioengineering}, \orgname{University of California San Diego}, \orgaddress{\street{9500 Gilman Dr}, \city{La Jolla}, \postcode{92093}, \state{CA}, \country{USA}}}

\affil[4]{\orgdiv{Department of Electrical and Electronic Engineering}, \orgname{Imperial College London}, \orgaddress{\street{Exhibition Rd}, \city{London}, \postcode{SW7 2AZ}, \country{UK}}}

\affil[5]{\orgdiv{International Centre for Neuromorphic Engineering}, \orgname{Western Sydney University, Penrith}, \orgaddress{\street{Second Ave }, \city{Kingswood}, \postcode{2747}, \state{NSW}, \country{Australia}}}

\affil[6]{\orgdiv{Biocomputation Group}, \orgname{University of Hertfordshire}, \orgaddress{\street{Exhibition Rd}, \city{London}, \postcode{SW7 2AZ}, \country{UK}}}

\affil[7]{\orgdiv{Department of Physics}, \orgname{University of California, Berkeley}, \orgaddress{\street{University Avenue and Oxford St}, \city{Berkeley}, \postcode{94720}, \state{CA}, \country{USA}}}

\affil[8]{\orgdiv{Redwood Center for Theoretical Neuroscience and Helen Wills Neuroscience Institute}, \orgname{University of California, Berkeley}, \orgaddress{\street{University Avenue and Oxford St}, \city{Berkeley}, \postcode{94720}, \state{CA}, \country{USA}}}

\affil[9]{\orgdiv{Chair of Highly-Parallel VLSI-Systems and Neuro-Microelectronics},\orgname{Technische Universit{\"a}t Dresden}, \orgaddress{\street{Mommsenstra{\ss}e 12}, \city{Dresden}, \postcode{01069}, \country{Germany}}}

\affil[10]{\orgdiv{Department of Electrical and Computer Engineering}, \orgname{University of California, Santa Cruz}, \orgaddress{\street{1156 High Street}, \city{Santa Cruz}, \postcode{95064}, \state{CA}, \country{USA}}}

\affil[11]{ \orgname{Politecnico di Torino}, \orgaddress{\street{Corso Duca degli Abruzzi, 24}, \city{Torino}, \postcode{10129}, \country{Italy}}}

\affil[12]{\orgdiv{Department of Electrical and computer engineering}, \orgname{Johns Hopkins University}, \orgaddress{\street{3400 N. Charles Street}, \city{Baltimore}, \postcode{21218}, \state{MD}, \country{USA}}}

\affil[13]{\orgname{SLAC National Accelerator Laboratory
}, \orgaddress{\street{2575 Sand Hill Road
}, \city{Menlo Park
}, \postcode{94025}, \state{CA}, \country{USA}}}

\affil[14]{\orgdiv{Materials Science and Engineering}, \orgname{Stanford University}, \orgaddress{\street{450 Jane Stanford Way}, \city{Stanford}, \postcode{94305}, \state{CA}, \country{USA}}}

\affil[15]{\orgname{Scads.AI: Center for scalable data analytics and artificial intelligence}, \orgaddress{\street{Strehlener Street 12, 14}, \city{Dresden}, \postcode{01069}, \country{Germany}}}

\abstract{We introduce NeuroSA, a neuromorphic architecture specifically designed to ensure asymptotic convergence to the ground state of an Ising problem using a Fowler-Nordheim quantum mechanical tunneling based threshold-annealing process. The core component of NeuroSA consists of a pair of asynchronous ON-OFF neurons, which effectively map classical simulated annealing dynamics onto a network of integrate-and-fire neurons. The threshold of each ON-OFF neuron pair is adaptively adjusted by an FN annealer and the resulting spiking dynamics replicates the optimal escape mechanism and convergence of SA, particularly at low-temperatures. To validate the effectiveness of our neuromorphic Ising machine, we systematically solved  benchmark combinatorial optimization problems such as  MAX-CUT and Max Independent Set. Across multiple runs, NeuroSA consistently generates distribution of solutions that are concentrated around the state-of-the-art results (within 99\%) { or surpass the current state-of-the-art solutions for Max Independent Set benchmarks}. Furthermore, NeuroSA is able to achieve these superior distributions without any graph-specific hyperparameter tuning. For practical illustration, we present results from an implementation of NeuroSA on the SpiNNaker2 platform, highlighting the feasibility of mapping our proposed architecture onto a standard neuromorphic accelerator platform. }


\maketitle

\section{Introduction}\label{sec1}

Quadratic Unconstrained Binary Optimization (QUBO) and Ising models are considered fundamental to solving many combinatorial optimization problems~\cite{Barahona1982Ising,Andrew2014IsingNP,Mohseni2022IsingCOP} and in literature, both classical and quantum hardware accelerators have been proposed to efficiently solve QUBO/Ising problems~\cite{hamerly2019QuantumVSCIM,tanahashi2019ClassicalIsing}. These accelerators use some form of annealing to guide the collective dynamics of the underlying optimization variables (e.g. spins) toward the global optima of the COP, which correspond to a specific system's ground energy states. Examples of such accelerators include superconducting qubit-based quantum annealers~\cite{King2022CoherentQA}, optical and digital coherent Ising machines (CIM) on~\cite{Mwamsojo2023OpticCIMCOP,Honjo2021DigitalCIM}, CMOS-based oscillator networks~\cite{Graber2023CMOSOSCNetwork, Maher2024CMOSVOISING}, memristor-based Hopfield Network~\cite{Cai2020PowerefficientCO,Jiang2023QuantumInspiredAnnealingMem, Fahimi2021weightAnnealingMem}, and digital circuit-based simulated annealing (SA)~\cite{Isakov2015OptimisedSA,kihara2017SAfpga,yamaoka2015SAcmos,okuyama2016approxSA}. Quantum Ising machines that use quantum annealing can theoretically guarantee finding the optimal solution to the QUBO/Ising problem, however, the approach cannot yet be physically scaled to solve large-scale problems~\cite{kadowaki1998QA,das2008QAanalog,farhi2001quantum,albash2018Qadiabatic}. On the other hand, classical QUBO / Ising solvers exploit nonlinear dynamics of the system to explore the solution space and avoid local minima. Simulated bifurcation machine (SBM)~\cite{Goto2019SBM} and probabilistic bit (p-bit)-based stochastic solvers relax the binary constraints and exploit the thermal noise inherent in magnetic tunnel junctions (MTJs) hardware~\cite{Giovanni2023MTJPIM,Singh2024CMOSMTJpim}. Memristor-based Hopfield Networks~\cite{Cai2020PowerefficientCO} also harness device-intrinsic noise and can achieve low energy-to-solution and time-to-solution metrics. However, if the objective is to produce solutions that consistently approach the state-of-the-art (SOTA) optimization metric or even exceed the SOTA (discover better solutions), then the precision of computation and the dynamic range of peripheral read-out circuits limit the performance of analog Ising machines. This is highlighted in Fig. 1a for three classes of COPs: low-complexity (L), medium-complexity (M), and high-complexity (H). Here, the complexity of the problems is characterized based on various metrics such as number of variables, graph density, and graph transitivity. For low-complexity problems, most solvers can find the SOTA (or the ground state) for every run. This results in a distribution that is concentrated around SOTA, as highlighted in Fig. 1a. However, for COPs with higher complexity, the distribution of solutions can exhibit a large variance, because: (a) the ground state could be significantly separated from the SOTA (shown in Fig. 1a); and (b) even finding an approximate solution could be NP-hard (or requires exponential time or high precision). Scaling analog Ising machines architecturally to larger problems would also require tiling small-sized memristor/analog crossbars with additional communication infrastructure (using spikes or digital encoding). Digital Ising architectures do not suffer from the precision and scaling challenges. Furthermore, in memristor-based Ising machines, the statistics of the device noise and its modulation have to be experimentally tuned/calibrated such that the results approach the state-of-the-art (SOTA) solution quality. There is no theoretical guarantee that such a machine would improve upon the SOTA or asymptotically converge to the ground state. 

\begin{figure}[h!]
\centering
\includegraphics[width=0.95\textwidth]{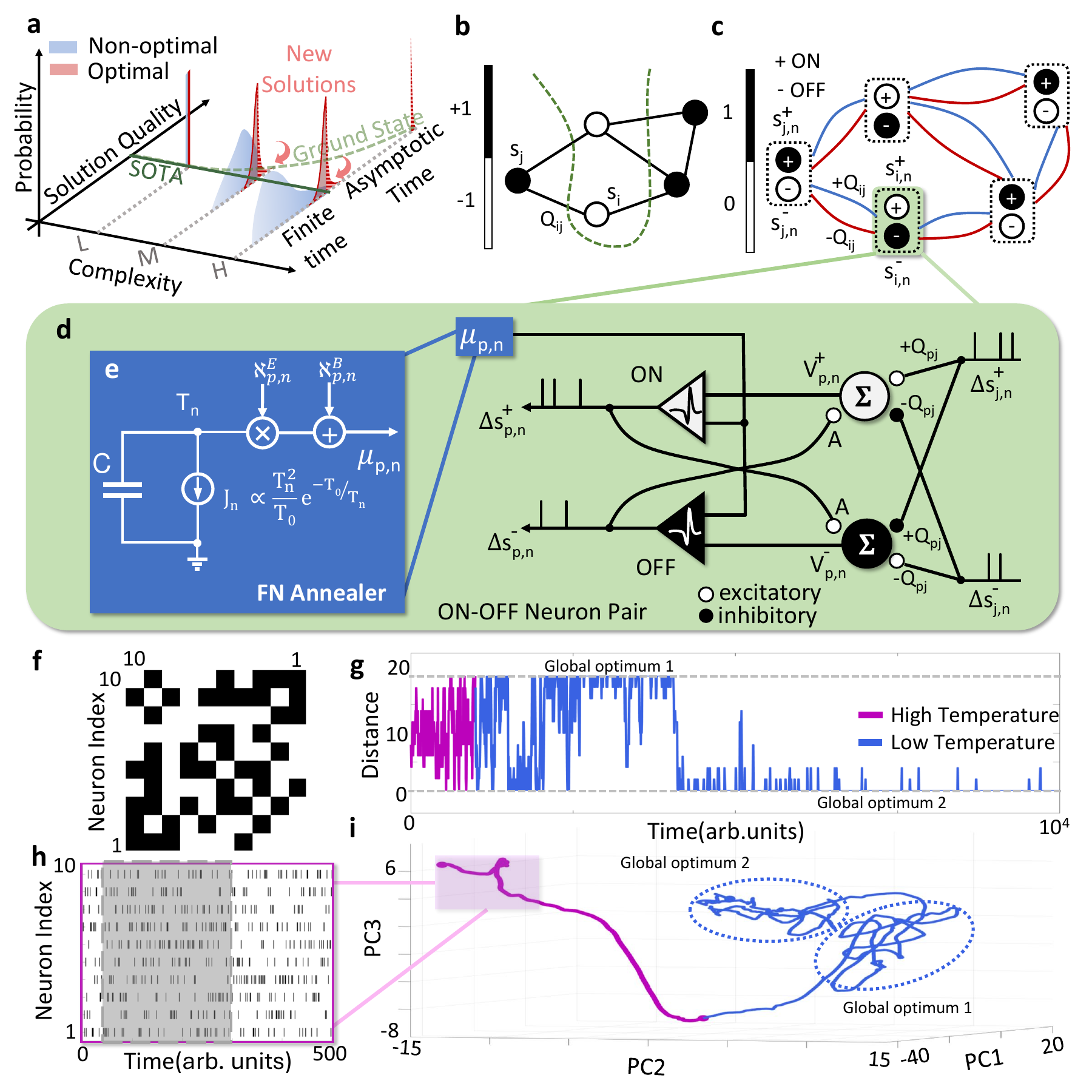}
\caption{NeuroSA motivation for mapping of optimal simulated annealing into a neuromorphic architecture: (a) {Illustration of the distribution of solutions generated by optimal and non-optimal Ising machines for different COP complexity: Low(L), Medium(M), and High(H). } {An ideal Ising/QUBO solver} produces distribution of solutions that is concentrated near the SOTA and has the potential to produce novel, previously unknown solution that is closer to the Ising ground state; A MAX-CUT problem defined over a (b) graph with weights $Q_{ij}$ which is decomposed into (c) pairs of ON-OFF neurons by NeuroSA. (d) Each ON-OFF integrate-and-fire neurons are coupled to each other by an excitatory synapse with weight $A$ and the pair is connected differentially to other ON-OFF neuron pairs through the synaptic weights $Q_{ij},-Q_{ij}$. The thresholds for both ON-OFF neurons are dynamically adjusted by an (e) FN annealer which comprises an FN integrator, an exponentially-distributed noise source $\mathcal{N}_n^E$ and a Bernoulli noise source $\mathcal{N}_n^B$; Illustration of NeuroSA dynamics for a MAX-CUT graph with 10 vertices connected by a weight matrix $\mathbf{Q}$ shown in (f); (g) Evolution of the distance between the solutions generated by NeuroSA to the two known ground state solutions at a given time-instant which highlights the escape mechanisms in the high- and low-temperature regimes; (h) Raster plot of aggregated spiking activity generated by the ON and OFF neuron pairs, and (i) visualization of the NeuroSA trapping and escape dynamics using a {Principle Component Analysis (PCA)}-based projection of the network spiking activity estimated within a moving time-window.}\label{nc_fig1}
\end{figure}

Advances in neuromorphic hardware have now reached a point where the platform can simulate networks comprising billions of spiking neurons and trillions of synapses. Implementations of these neuromorphic supercomputers range from commercial-off-the-shelf CPU-, and GPU-based platforms~\cite{kang2020cpuNeuro1,ivanov2022cpuNeuro2} to custom FPGA-, multi-core-, and ASIC-based platforms~\cite{furber2014spinnaker,Park2017Hiaer,davies2018loihi}. As an example, the SpiNNaker2 microchip~\cite{mayr2019spinnaker2}, which has been used for some experiments in this paper, can integrate more than 152,000 programmable neurons with more than 152 million synapses in total. Although the primary motivation for developing these platforms has been to emulate/study neurobiological functions~\cite{van2018spinnSim, rhodes2020corticalSim} and to implement artificial intelligence (AI) tasks~\cite{liu2018memorySpinnaker,lin2018computeLoihi,gonzalez2024spinnaker2} with much needed energy efficiency~\cite{shankar2023energy}, it has been argued that the neuromorphic advantage can be demonstrated for tasks that can exploit noise and nonlinear dynamics inherent in current neuromorphic systems. These approaches exploit the emergent properties of an energy or entropy minimization process~\cite{hopfield1982hpf,friston2010freeEnergyBrain}, phase transitions and criticality~\cite{chialvo2010phase}, chaos~\cite{sompolinsky1988chaos}, and stochasticity~\cite{mcdonnell2011noise1,schneidman1998channelNoise,branco2008DentriteNoise}, both at the level of an individual neurons~\cite{jaeger2004harnessingNonlinear, hoppensteadt1999oscillatory,Goto2019SBM} and at the system level such as the Hopfield Network~\cite{hopfield1982hpf,Cai2020PowerefficientCO} and Boltzmann machines~\cite{fischer2012RBoltzmann,amin2018QBoltzmann}. Recently, neuromorphic advantages in energy efficiency have been demonstrated for solving optimization problems~\cite{lin2018computeLoihi,Mehta2022FNDAM} and for simulating stochastic systems like random walks~\cite{Smith2022NeuromorphicRandomWalk}. These specific implementations exploit the high degree of parallelism inherent in neuromorphic architectures for efficient Monte Carlo sampling, and for implementing Markov processes both of which are important for solving Ising problems. {Other promising approach has been to use neuromorphic architectures to solve semi-definite programming (SDP), which is an approximation method to solve many COPs~\cite{ragha2008}. In fact, under the correctness assumption of the unique games conjecture~\cite{khot2010}, SDPs have been employed to produce optimal solutions amongst all polynomial-time algorithms~\cite{ragha2008}. Examples of such implementations include the mapping of the Goemans-Williams (GW) SDP onto Intel Loihi ~\cite{Aimone2023GWLoihi,Aimone2022StochasticNeuronMaxcut} for MAX-CUT COPs. However, GW and other SDP algorithms only provide lower bounds on the solution quality and not on the variance of distribution of solutions. Furthermore, the precision required to solve the SDP algorithm could be high to be able to achieve the SOTA. As a result, even achieving the SDP approximation quality bound precisely or exceeding the bound (for specific graphs and runs) still requires a brute-force (exponential or sub-exponential) search around the SOTA. Also note that for COPs like MAX-CUT, even an improvement by a single cut could represent a significant step, as the improvement could only be achieved by finding a novel solution.}

Simulated Annealing (SA) algorithms, on the other hand, can provide guarantees of finding the QUBO/Ising ground state, provided the annealing schedule follows specific dynamics~\cite{Kirkpatrick1983,geman1984stochastic,Hajek1988OptimalCooling}. Previous attempts to solve Ising problems using neuromorphic hardware~\cite{Fahimi2021weightAnnealingMem,Jiang2023QuantumInspiredAnnealingMem} emulated SA dynamics, which did not guarantee asymptotic convergence to the Ising ground state.   Hence, a neuromorphic architecture that is functionally isomorphic to the SA algorithm with an optimal annealing schedule should produce high-quality solutions across different runs. This feature is highlighted in Fig. 1a by the desired (or optimal) distribution that is concentrated near the SOTA. Furthermore, as shown in Fig. 1a, having an asymptotic optimality guarantee will also ensure that a long run-time might produce a solution that is at least better than the current SOTA, if the current SOTA is not already the ground state of the COP. However, the dynamics of the SA algorithm can be very slow which motivates its mapping onto  a neuromorphic architecture and accelerator. 

How can optimal simulated annealing algorithms be mapped onto large-scale neuromorphic architectures? The key underpinnings of any neuromorphic architecture are: (a) asynchronous (or Poisson) dynamics that are generated by a network of spiking neurons; and (b) efficient and parallel routing of spikes/events between neurons across large networks. Both these features are essential for solving the Ising problem and efficient mapping of SA onto neuromorphic architecture. In its general form 
Ising problem minimizes a function (or a Hamiltonian) $H(\mathbf{s})$ of the spin state vector $\mathbf{s}$ according to
\begin{equation}
    \min_{\mathbf{s}\in\{-1, +1\}^{D}}H(\mathbf{s})=\frac{1}{2}\mathbf{s}^\intercal\mathbf{Q}\mathbf{s} + \mathbf{b}^T\mathbf{s}
    \label{eq_intro_ising_energy}
\end{equation}
where $\mathbf{b} \in \mathbb{R}^D$ represents an external field or bias vector and can also be used to introduce additional constraints into the Ising problem~\cite{Lucas2014Ising}. In SI Sections S1, 2, and 3 we consider these specific cases, but for the ease of exposition, our focus will be on problems where $\mathbf{b} = \mathbf{0}$. In such cases the Ising problem becomes equivalent to the MAX-CUT problem (shown in SI Section S1 and S2), which is easy to visualize and analyze~\cite{Goemans95maxcut}.  
For a simple MAX-CUT graph depicted in Fig. 1b, each of the spin variables, denoted as $s_i \in \{-1,+1\}$, where $i = 1,.., D$, is associated with one of the $D$ vertices in the graph $\mathcal{G}$. The graph's edges are represented by a matrix $\mathbf{Q}\in \mathbb{R}^{D\times D}$, wherein $Q_{ij}$ signifies the weight associated with the edge connecting vertices $i$ and $j$. Given the graph $\mathcal{G}$, the objective of the MAX-CUT problem is to partition the vertices into two classes, maximizing the number of edges between them. If an ideal asynchronous operation is assumed (see Methods section~\ref{sec_methods_async_ising}), then at any time instant $n$, only one spin (say the $p^{th}$ spin) changes its state by $\Delta\mathbf{s}_{p,n} \in \{-1,0,+1\}$. In this case, the function $H$ decreases or $\Delta H_n < 0$, if and only if the condition
\begin{equation}
    \Delta s_{p,n}\left[\sum_{j=1}^DQ_{pj}s_{j,n-1}\right] < 0
    \label{eq_intro_dH_async}
\end{equation}
is satisfied. {The inherent parallelism of neuromorphic hardware ensures that the pseudo-gradient $\sum_{j=1}^DQ_{pj}s_{j,n-1}$ is computed at a rate faster than the rate at which events $\Delta s_{p,n}$ are generated.} The condition described in Eq.~\ref{eq_intro_dH_async}, when combined with the simulated annealing's probabilistic acceptance criterion~\cite{Kirkpatrick1983}, leads to a neuromorphic mapping based on coupled ON-OFF integrate-and-fire neurons where the $p^{th}$ ON-OFF pair is shown in Fig. 1d. Please refer to the Methods section~\ref{sec3_methods} for the derivation of the ON-OFF neuron model. The $p^{th}$ ON-OFF neuron pair is differentially connected to the $j^{th}$ neuron pair through the synaptic weights $Q_{pj},-Q_{pj}$. The spikes generated by this $p^{th}$ post-synaptic ON-OFF neuron pair  $\Delta s_{p,n}^+, \Delta s_{p,n}^- \in \{0,1\} $ differentially encodes the change in the $p^{th}$ spin state, and the cumulative state $s_{j,n}$ of the pre-synaptic neuron is estimated by continuously integrating the input spikes $\Delta s_{j,n}^+, \Delta s_{j,n}^-$ received from the $j^{th}$ neuron. To ensure that the spiking activity of the ON-OFF neuronal network is functionally isomorphic to the acceptance/rejection dynamics of an SA algorithm, the firing threshold $\mu_{p,n}$ of the $p^{th}$ neuron adjusted over time by a Fowler-Nordheim (FN) annealer, shown in Fig. 1d.  

We employ a Fowler-Nordheim dynamical system can produce dynamic thresholds that correspond to the SA optimal annealing schedule.
One of the key results from the SA literature~\cite{geman1984stochastic, Hajek1988OptimalCooling} is the proposition that a temperature cooling schedule that follows $\sim\frac{c}{\log(1+n)}$ can guarantee asymptotic convergence to the QUBO/Ising ground state, where $c$ denotes the largest depth of any local minimum of Ising Hamiltonian $H(\mathbf{s})$ {and $n$ denotes the discrete time step}.  A dynamical systems model in Fig. 1e comprising of a time-varying Fowler-Nordheim (FN) current element $J_n$~\cite{Fowler1928Tunn} can generate the optimal $T_n=\frac{T_0}{\log(1+n/C)}$ according to~\cite{Zhou2017FNtimer}, where $T_0$ and $C$ are Fowler-Nordheim annealing hyperparameters (see Methods section~\ref{sec_methods_FN_annealer}). 
{The FN dynamics can be either be generated using a physical FN-tunneling device~\cite{Zhou2017FNtimer} or can be emulated by implementing a FN-tunneling dynamical systems model on a digital hardware. Here, we chose the digital emulation because of the precision required for SA in the low-temperature regime.} 
The FN dynamics can then be combined with the independent identically distributed (i.i.d) random variables $\mathcal{N}_{p,n}^E$ and $\mathcal{N}_{p,n}^B$ to determine the dynamic firing threshold $\mu_{n,p}$ for each ON-OFF integrate-and-fire neuron pair $p$. $\mathcal{N}_{p,n}^E$ is drawn from an exponential distribution where as $\mathcal{N}_{p,n}^B$ is drawn from a Bernoulli distribution with values $\{0,1\}$. The choice of the two distributions ensures that every neuron has a finite probability of firing, which is equivalent to satisfying the irreducibility and aperiodicity conditions in SA.

The ON-OFF neuron pair and the integrated FN annealer, shown in Fig. 1c form the basic computational unit of NeuroSA which can be used to solve Ising problems on different neuromorphic hardware platforms. Due to the functional isomorphism between NeuroSA and the optimal SA algorithm, the hardware could accelerate and asymptotically approach the Ising ground state, as highlighted in Fig. 1a. We show in the results that even when NeuroSA solves a COP for a finite duration, the machine produces the distribution of solutions that are concentrated around the SOTA, as shown in Fig. 1a, and this is achieved without significant tuning of hyperparameters.

\section{Results}\label{sec2}
{We first demonstrate the dynamics of the NeuroSA architecture for a small MAX-CUT problem, where the optimization landscape and the search dynamics can be easily visualized and where the Ising ground states can be determined using brute-force search. We then evaluate NeuroSA on larger-scale benchmark COPs for which the SOTA solution quality is well documented in the literature~\cite{medium2019BenchmarkingMAXCUT,yik2024neurobenchframeworkbenchmarkingneuromorphic}.}

\subsection{Experiments using small-scale graphs}\label{sec_res_10_dynamics}
NeuroSA is first applied to a 10-node MAX-CUT graph described by the interconnection matrix $\mathbf{Q}$ shown in Fig. 1f. For this graph, the two degenerate ground states exist (due to the gauge symmetry where $\mathbf{s} \leftrightarrow -\mathbf{s}$) and can be found by an exhaustive search over all possible spin state configurations. Fig. 1g plots the Manhattan distance measured with respect to one of the ground states. For this specific example, there exist two ground states (due to gauge symmetry) which are labeled as global optimum 1 and 2 in Fig. 1g. It is evident from Fig. 1g that, like SA, the dynamics of NeuroSA can be categorized into two phases: the high-temperature regime and the low-temperature regime. These regimes are determined by the firing threshold $\mu(t)$ in~Fig. 1d. To visualize the network dynamics, the aggregated spiking rate for each of the ON-OFF neurons is calculated using a moving window as shown in Fig. 1h. The population firing dynamics are projected onto a reduced $3$-D space spanned by its three most dominant principal component vectors, as shown in Fig. 1i. This Principal Component Analysis (PCA)-based approach is a standard practice in analyzing spiking data~\cite{stopfer2003intensity} and details of the approach are described in Methods Section~\ref{sec_methods_PCA}. In the high-temperature phase, the network dynamics evolve along the network gradient, resulting in faster convergence along a smooth trajectory as depicted in Fig. 1i. As the temperature cools down, NeuroSA exploration is trapped in the neighborhood of one of the two global optima, as depicted in Fig. 1g and i. In the neighborhood, the dynamics follow a random walk; however, the dynamics can periodically escape one of the global optima for further exploration and possibly converge to the second global optimum. This is highlighted in Fig. 1i by the continuous trajectory connecting the two optima/attractors. Eventually, as shown in Fig. 1g, due to the annealing process the dwell-time of dynamics in the neighborhood of the optima increases over time. Note that state-space exploration in the low-temperature regime is a significant problem in SA algorithms, and heuristics such as hybrid quantum-classical methods~\cite{Layden2023MCMC} have been proposed to accelerate this process. In NeuroSA, the Fowler-Nordheim dynamical process allows for a finite probability of escape even in the low-temperature regime. 

\begin{figure}[h]
\centering
\includegraphics[width=\textwidth]{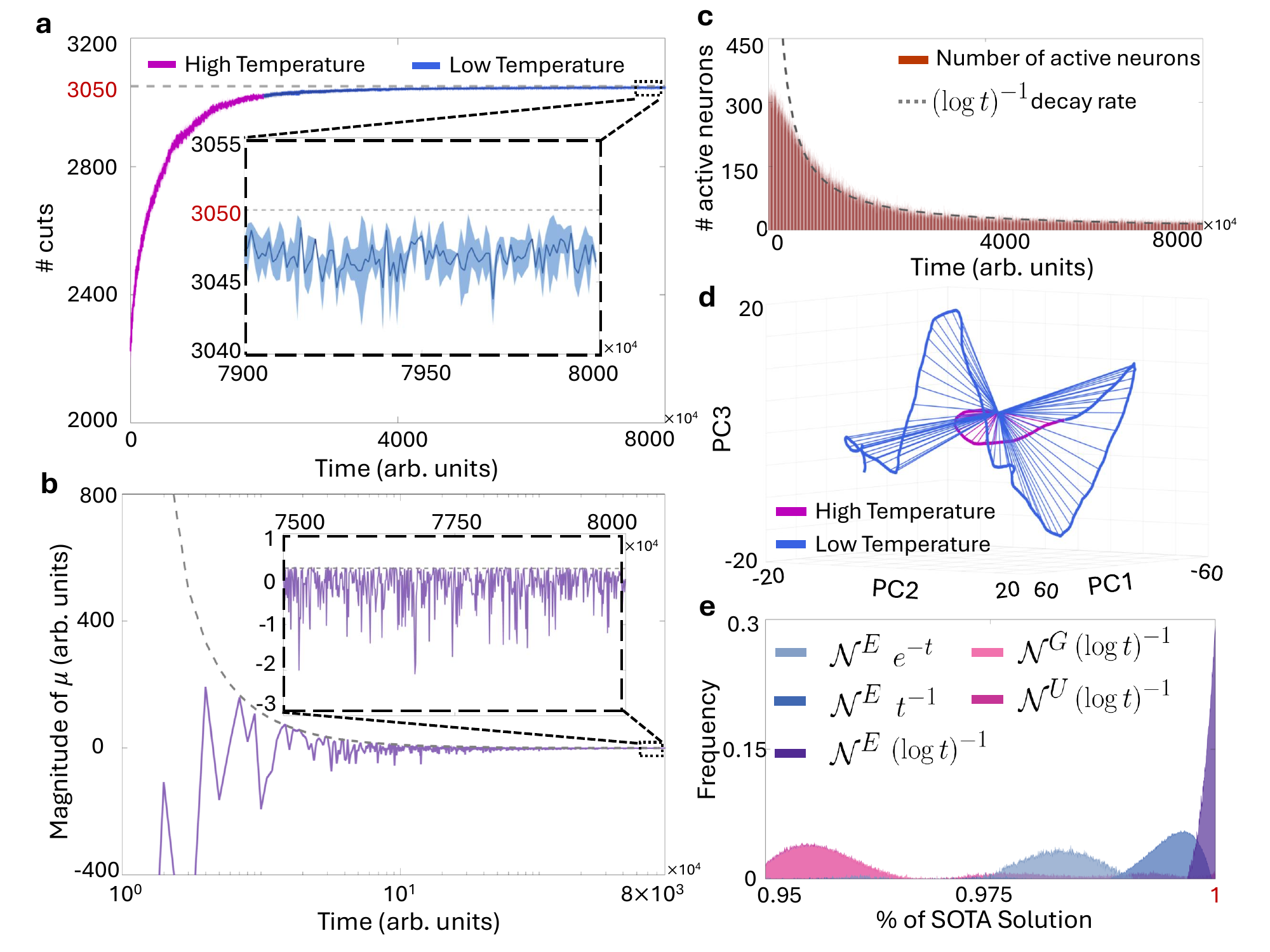}
\caption{NeuroSA dynamics for the $G15$ MAX-CUT graph with 800 vertices and 4661 edges: (a) convergence plot showing steady increase in the solution quality with the inset showing fluctuations near $3050$ cuts which is the current SOTA for this graph; (b) dynamics of the firing threshold with inset showing sparse but large fluctuations that trigger escape mechanisms; (c) plot showing the number of active neurons decaying following $\sim\frac{1}{\log t}$ without the contribution of the Bernoulli r.v. $\mathcal{N}^B$; (d) PCA trajectory of the NeuroSA dynamics where the initial (high temperature) regime follows a path defined by the network gradient and the trajectory near convergence (or low-temperature path) exhibits expanding exploration of the solution space; and (e) distribution of the $G15$ solutions obtained for different annealing schedules ($e^{-t},(\log t)^{-1}, t^{-1}$) and noise statistics (exponential - denoted by $\mathcal{N}^E$, Gaussian - denoted by $\mathcal{N}^G$ and Uniform - denoted by $\mathcal{N}^U$)}\label{nc_fig2}
\end{figure}

\subsection{Experiments using medium-scale graph}\label{sec_res_g15_dynamics}

We next apply NeuroSA to a MAX-CUT problem on a graph where the ground state is not known, but the SOTA solution is well documented. We chose the G$15$, an $800$-node, binary weighted, planar graph~\cite{Gset} for which the SOTA is $3050$ cuts~\cite{Goto2019SBM}. The NeuroSA architecture was simulated on a CPU-based platform and the hardware mapping procedure is described in the Methods Section~\ref{sec_methods_sync_impl} and the pseudo-code for the implementation is presented in the SI Section S4. Fig. 2a shows the solution found by NeuroSA over time converging to the SOTA states.

The dynamics of the noisy firing threshold $\mu_n$ is shown in Fig. 2b and is bounded by $T_n\sim\frac{1}{\log(n)}$ (depicted in the figure as the dotted line), produced by the FN annealer. As the envelope of the threshold decreases, it inhibits the firing probability of neurons as shown by the histogram in Fig. 2c.  During the initial phases of convergence, there is a gap between the $\frac{1}{\log(n)}$ envelope and the number of active neurons (or the neurons whose membrane potential exceeds the firing threshold). Thus, in this phase, the dynamics of the network seems to be governed by the network gradient. However, the tails of the histogram fit the $\frac{1}{\log(n)}$ dynamics reasonably well, highlighting the influence of the FN-based escape mechanism on the network dynamics. The network population dynamics are depicted by the PCA trajectory of the aggregated spiking rate and are plotted in Fig. 2d. Similar to the results for the small-scale graph in Fig. 1i, the trajectory reflects the evolution of the NeuroSA system as it explores the solution space. The exploration in the high temperature regime follows a more confined trajectory in the PCA space, indicating that the network dynamics evolve in the direction of the Ising energy gradient. As the temperature cools down near convergence, the NeuroSA dynamics are dominated by the random walk and sporadic escape mechanisms with no specific direction, as shown in Fig. 2d. 

Fig. 2e plots the distribution of the solution quality (normalized with respect to SOTA) when different cooling schedules are chosen and the r.v. $N^E$ are chosen from different statistical distributions. In particular, previous neuromorphic Ising models and stochastic models have used Gaussian noise as a mechanism for asymptotic exploration and for escaping local minima. However, the results in Fig. 2e show that this approach produces distributions with longer tails and in some cases solutions that are significantly worse than the SOTA. Only for an FN annealer and an exponentially distributed noise $N^E$, the distribution of solutions obtained by NeuroSA is more concentrated around the SOTA. 

\subsection{Benchmarking NeuroSA for different MAX-CUT graphs}\label{sec_res_sweep}

Next the NeuroSA architecture was benchmarked for solving MAX-CUT problems on different Gset graphs. Fig. 3 provides a detailed evaluation of the NeuroSA algorithm's performance on the Gset benchmarks, with results generated using both traditional CPU-based and the SpiNNaker2 platform. The architecture is configured similarly for both hardware platforms and across all benchmark tests. This uniformity is important, as it demonstrates that NeuroSA's performance is robust across and agnostic to different MAX-CUT graph complexities. Also, it obviates painstaking hyperparameter tuning for each set of graphs or problems.

\begin{figure}[ht!]
\centering
\includegraphics[width=0.95\textwidth]{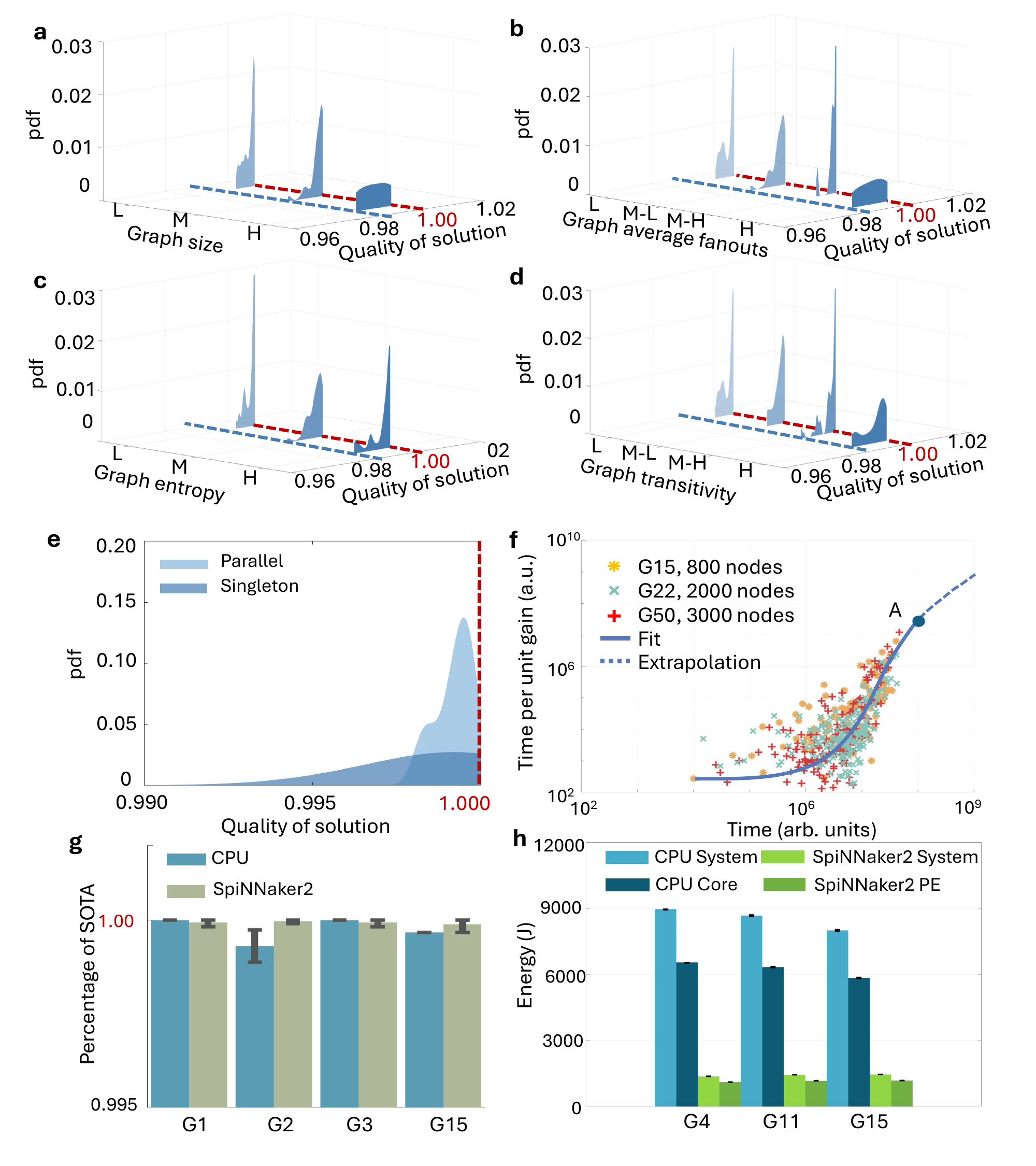}
\caption{{NeuroSA results for MAX-CUT and MIS benchmarks. (a)-(d) Empirical probability density functions (pdf) of the solutions on Gset benchmarks~\cite{medium2019BenchmarkingMAXCUT}. The solutions fall within the interval $(0.989, 1.00)$ of SOTA, indicated by the red and blue dotted lines. The results are ordered with ascending complexity: Low(L), Medium-Low(M-L), Medium(M), Medium-High(M-H), High(H). Complexity metrics: (a) the number of graph vertices, L=\{$800, 1,000$\}, M=\{$2,000, 3,000$\} and H=\{$5,000, 10,000$\}; (b) the average fan-out, where L=$2.0$, M-L=$6.0$, M-H=$10.0$, H=$24.0$; (c) the graph entropy, where L=[0, 2), M=(2, 4) and H=(4, 5); (d) the graph transitivity, where L=[0, 0.001], M-L=(0.001, 0.05], M-H=(0.05, 0.14), H=[0.14, 0.16). (e) Parallel search comprising of 5 NeuroSA instances yields more consistent results than a single search running for 5$\times$ the duration of the parallel search. (f) Instantaneous time per unit gain in solution for $3$ different Gset benchmarks approaches an exponential/sub-exponential run-time. (g) Results from NeuroSA implemented on the SpiNNaker2 neuromorphic platform, where error bars plots the SD. (h) The energy-to-solution results on both CPU and SpiNNaker2 platforms, where the error bar plots the maximum and minimum energy dissipation across runs.
} 
}\label{nc_fig3}
\end{figure}

Fig. 3a-d report the distribution of solutions obtained by NeuroSA for different Gset MAX-CUT benchmarks. For each NeuroSA run, the solution (number of cuts) is normalized with respect to the SOTA solution which is highlighted in Fig. 3a-d as the unity marker. Note that for MAX-CUT problems, the SOTA is from previously reported results in literature~\cite{medium2019BenchmarkingMAXCUT,Goto2019SBM}. The probability density function is empirically obtained as described in Methods Section~\ref{sec_methods_pdf}. While determining the complexity of COPs is in itself a difficult task in itself, to gain more insight, the distributions are organized based on commonly used graph complexity metrics. Fig. 3a-d organizes the {results} by: (a) the graph size, namely the number of vertices; (b) the average fan-out per node;  (c) the graph entropy, measuring the randomness in connectivity~\cite{dehmer2011graphEntropy}; and (d) the network transitivity, focusing on node connectivity density, indicating clustering within the network~\cite{holland1971transitivity}. The average fan-out per node measures the typical number of direct connections (outgoing edges) each node has, providing a basic indication of the graph's overall connectivity and potential for information spread. Graph entropy, on the other hand, quantifies the randomness or disorder in the graph by analyzing the distribution of these connections across all nodes. It offers insights into how evenly or unevenly the connections are distributed, with higher entropy indicating a more complex or disordered network structure. The global clustering coefficient, also known as transitivity, measures the degree to which nodes in a graph tend to cluster together. This coefficient assesses the overall tendency of nodes to create tightly knit groups, with higher values suggesting a greater prevalence of interconnected triples of nodes, which can indicate a robust local structure within the network. While the average fan-out per node provides a simple measure of connectivity, it does not capture the nuances of how these connections are configured, which is where graph entropy and the global clustering coefficient come into play. Graph entropy complements the average fan-out by assessing the variability in node connectivity, highlighting potential inequalities or irregularities in how nodes are linked. In contrast, the global clustering coefficient focuses on the tendency to form local groups, offering a view of the graph's compactness and the likelihood of forming tightly connected communities.  Together, these metrics provide a multi-dimensional view of a graph's complexity, indicating not only how many connections exist, but also how they are organized and how they foster community structure and network resilience.

The results show that the NeuroSA solutions consistently reach within $99\%$ of the SOTA, despite of the complexity of the graph. However, as expected, the variance of the distribution does increase with complexity which indicates that one can also speculate on the general complexity of the COP itself. In Fig. 3e we show that there exists an algorithmic advantage in mapping NeuroSA onto a parallel architecture, like the SpiNNaker2 platform. We compare the results obtained by fice NeuroSA instances launched in parallel, where each instance executes for $10^8$ iterations against a singleton NeuroSA instance that runs for $5 \times 10^8$ iterations. We then aggregate the results from the five parallel instances by taking the average, and plot the probability distribution of the quality of solution for both parallel and singleton instances. As shown in Fig. 3e, the parallel search results in a more concentrated distribution around the SOTA solution than the long-running singleton approach. Effectively, the parallel NeuroSA instances take different trajectories to the solution and the aggregation of multiple searches results in amplifying the probability of reaching the neighborhood of the SOTA/ground state solution within a finite time constraint. However, we would like to point out that the optimal annealing schedule promises global optimality only in the asymptotic (exponential) time limit. Therefore, improvements on the current solution to reach the ground state using a parallel approach would still rely on long run-time of one of many singleton NeuroSA instances. We extended the execution time by 1,000$\times$, to $10^{11}$ iterations to see if NeuroSA could find a novel solution for a few sets of MAX-CUT problems. Unfortunately, within finite run-time (both on CPU and SpiNNaker2), NeuroSA was unable to discover solutions that exceeded the SOTA for the Gset/MAX-CUT benchmarks.

To understand the difficulty in finding new solutions, Fig. 3f presents an analysis of the time required per unit gain in the solution for three distinct Gset benchmarks. The plots reveal an increasing cost (computational time) to achieve marginal improvement in the quality of the solution. The key metric in the plot Fig. 3f is the ratio between the time needed to obtain a unit improvement to the total run time. As it becomes harder to find better solutions, the ratio tends to unity (run-time becomes exponential or sub-exponential) This is shown by the extrapolation curve and the transition point $A$, which might highlight the point of diminishing returns. This could be taken as a hardware-agnostic stopping criterion (polynomial run-time) for a COP.

Fig. 3g shows the results when NeuroSA is implemented on the SpiNNaker2 platform (details provided in SI Section S5) for some of the Gset benchmarks. The results show similar or better solutions than the CPU/software implementation of NeuroSA. Note that NeuroSA is a randomized algorithm so each hardware implementation will exhibit different variance based on the quality of noise generator in the FN-annealer. The performance advantage achieved by mapping NeuroSA onto SpiNNaker2 is summarized in SI Section S5.2 and in Fig. 3h which show the energy-to-solution comparison with respect to the CPU-based implementation. The comparison is shown for three different 800-node MAX-CUT problem, and we compare the energy dissipated by the system (board-level) and by the processing-element-level (PE). Due to the distributed and multi-core nature of the SpiNNaker2 system, the PE energy also accounts for the inter-core communication consumption over the NoC. As shown in Fig. 3h, in either case, the SpiNNaker2 implementation outperforms the CPU-based implementation in terms of energy-to-solution for the same workload. Furthermore, this performance advantage is evident even if the current SpiNNaker2 implementation of NeuroSA is not fully optimized. 

\subsection{Benchmarking NeuroSA for MIS problems}
\begin{figure}[ht!]
\centering
\includegraphics[width=0.98\textwidth]{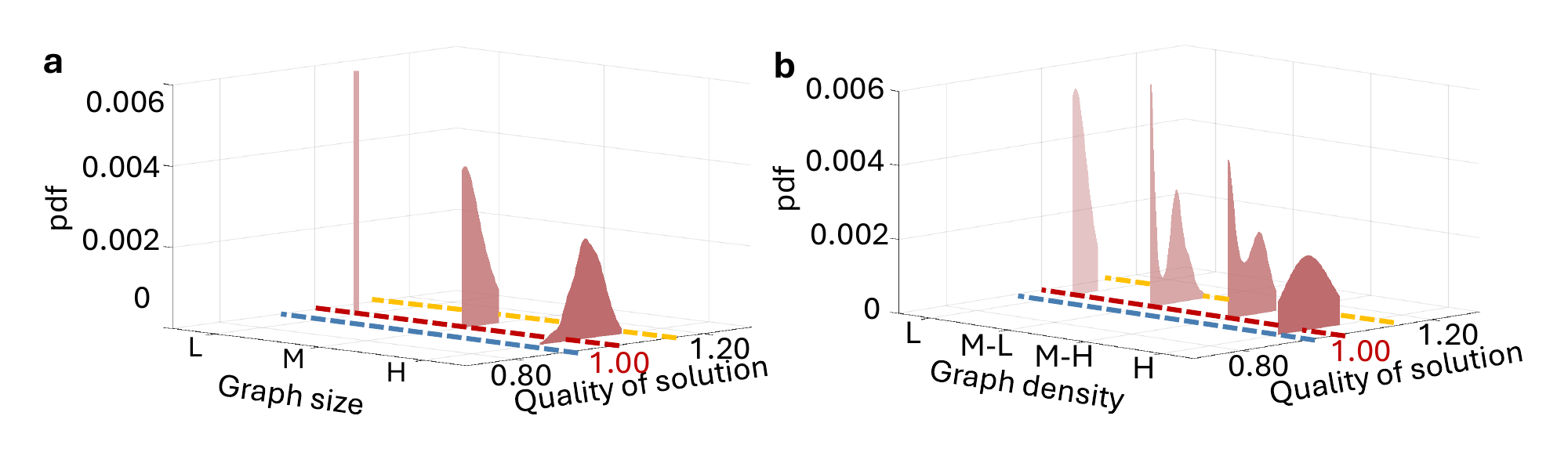}
\caption{{
NeuroSA results for MIS benchmarks showing that all solutions fall within the interval $(0.944, 1.11)$ of the SOTA (marked by dotted lines). Solutions are ordered according to complexity metrics: (a) number of graph vertices, with L=\{$10, 25, 50$\}, M=\{$100, 250, 500$\} and H=\{$1,000, 2,500, 5,000$\}; and (b) graph density, with L=$0.01$, M-L=$0.05$, M-H=$0.1$, H=$0.25$.
} 
}\label{nc_fig4}
\end{figure}
For the next set of experiments, the performance of NeuroSA architecture was evaluated for solving the maximum-independent-set (MIS) COPs. The problem involves searching for the largest set of vertices in a graph such that no two vertices in that set share an edge. Since the solutions to the MIS problem are subject to constraints, the problem is inherently harder than MAX-CUT where every state configuration is considered a valid solution. Details of mapping MIS COP onto an Ising formulation which can then be solved using NeuroSA are provided in SI Section S3. We evaluated NeuroSA on various MIS graphs from the NeuroBench benchmark suite~\cite{yik2024neurobenchframeworkbenchmarkingneuromorphic} and we characterized the distribution of the solutions generated for different graph sizes and graph densities, as shown in Fig. 4. Graph density represents the density of connections among vertices which controls the dimensionality of the system's state space and the effective dimensionality of the Ising energy landscape. Furthermore, higher density also alters the eigenvalue spectrum of the Hamiltonian, broadening it, and creating a richer set of critical points on the Ising energy surface. Similar to the MAX-CUT results, the MIS results also produce distributions are concentrated, without any hyperparameter tuning. However, unlike the MAX-CUT benchmarks, NeuroSA consistently finds solutions that are better than the current SOTA for MIS benchmarks. Also, note that the distributions in Fig. 4 are bounded below by $95\%$ of the SOTA, and this deviation occurs when the optimal set size just changes by $1$. Even though the variance of the distribution becomes larger with increasing complexity, a significant portion of the solutions surpasses the current SOTA, which validates the objective of the NeuroSA in Fig. 1a.   

\section{Discussions}\label{sec_discussion}
In this paper, we proposed a neuromorphic architecture called NeuroSA that is functionally isomorphic to a simulated annealing optimization engine. The isomorphism allows mapping optimal SA algorithms to neuromorphic architectures, providing theoretical guarantees of asymptotic convergence to the Ising ground state. The core computational element of NeuroSA is formed by an ON-OFF integrate-and-fire neuron pair that can be implemented on any standard neuromorphic hardware. Hence, NeuroSA can exploit the computational power of both existing and upcoming large-scale neuromorphic platforms, such as
SpiNNaker2 and HiAER-Spike. Inside each ON-OFF neuron pair is an annealer whose stochastic properties are dictated by a Fowler-Nordheim (FN) dynamical system. Collectively, the neuron model and the FN annealer generate population activity that emulates the sequential acceptance and rejection dynamics of the SA algorithm. 

The functional isomorphism between NeuroSA and the optimal SA algorithm also enables us to draw insights from SA dynamics to understand the emergent neurodynamics of NeuroSA and its convergence properties to a steady-state solution. For instance, the Bernoulli r.v. $\mathcal{N}^B_n$ within the FN annealer ensures the asynchronous firing such that only one of the neurons in NeuroSA fires at any given moment. From the perspective of SA, this asynchronous decomposition ensures that each combinatorial step of the COP is tractable, as described by the mathematical condition in Eq.~\ref{eq_methods_dH_async}. Also, the use of i.i.d Bernoulli r.vs in each ON-OFF neuron pair ensures that any pair can potentially fire (if its firing criterion is met), which in turn ensures that NeuroSA satisfies a key ergodic convergence criterion similar to that of SA algorithms. According to this criterion, every potential Ising state is reachable~\cite{Kirkpatrick1983}. The exponentially distributed r.v. $\mathcal{N}^E_n$ in the FN annealer upholds that an equivalent detailed balance criterion in SA~\cite{Kirkpatrick1983} is satisfied, thereby ensuring that the NeuroSA network attains an asymptotic steady-state firing pattern. This steady-state pattern corresponds to different mechanisms of exploring the Ising energy states, as depicted by the PCA network trajectories in Fig. 1i and Fig. 2d, with the assumption that the exploration will asymptotically terminate near the Ising ground state. This asymptotic convergence is guaranteed by modulating the dynamic threshold $\mu_n$ in the ON-OFF neuron pair, mimicking the optimal $\mathcal{O}(1/\log)$ temperature schedule in SA proposed by~\cite{geman1984stochastic,Hajek1988OptimalCooling}. Any choice of distribution other than the exponential distribution for the r.v. $\mathcal{N}^E_n$ will violate the SA's detailed balance criterion and hence the network might not encode a steady-state distribution.

In this work, two families of COPs, MAX-CUT and MIS, have been selected as COP benchmarks because they are well-studied COPs and the SOTA results for different MAX-CUT graphs are well documented in literature~\cite{Ozaeta_2022ExpectedQAOAGset}. As shown in the Results section~\ref{sec_res_sweep}, the NeuroSA architecture can consistently find solutions that are closer than 99\% SOTA metrics for different MAX-CUT benchmarks {and for the recently proposed MIS benchmark suite, NeuroSA improves on the current SOTA.} Note that the ground state solution for most of these graphs is still not known. In this regard, the asymptotic convergence to the ground state offered by the NeuroSA architecture is important as it can ensure good quality solutions across different runs, as highlighted in Fig. 3a-d. It's important to note that the $\mathcal{O}(1/\log)$ annealing schedule could make the convergence significantly slow which can be mitigated by sheer hardware acceleration offered by current and next-generation neuromorphic platforms. Also note that in Fig. 3a-d, the solutions obtained for some MAX-CUT graphs are inferior (percentage relative to the SOTA) compared to others, irrespective of the problem size (number of spin variables or ON-OFF neurons). This is because the problem complexity of some of the MAX-CUT problems is higher which implies that the NeuroSA architecture has to explore different regions of the energy landscape. By increasing the simulation run time and choosing a larger value of the hyperparameter $T_0$, the quality of the solution can be improved for all MAX-CUT graphs. The NeuroSA Ising machine can be used to solve other COP-like Hamiltonian path problems or Boolean satisfiability problems as well by optimizing a similar form of $H(\mathbf{s})$ in Eq.~\ref{eq_methods_ising_energy} but with real-valued $Q_{ij}$~\cite{tanahashi2019real_val_ising}. However, the energy landscape of the resulting $H(\mathbf{s})$ is more complex and hence would require different choices of $T_0$ and the simulation time to achieve SOTA solutions.

The NeuroSA architecture relies on the asynchronous nature of the SA acceptance dynamics which is directly encoded by spikes. The underlying assumption is that the spike from the neuron is propagated to all its synaptic neighbors before any other neuron in the network spikes. As shown by Eq.~\ref{eq_methods_firing_criteria_on} and~\ref{eq_methods_firing_criteria_off}, spike propagation from the ON-OFF neuron pair is equivalent to propagating pseudo-gradients in an SA algorithm. Most large-scale neuromorphic platforms rely on event routing mechanisms like Address event routing to transmit spikes across the network which incurs latency. As a result, if the spiking rate (equivalently the rate of the number of acceptances) is high, the asynchronous criterion specified in Eq.~\ref{eq_methods_dH_async} might not be satisfied. Furthermore, in practice, spikes (or event packets) might not be properly routed to the neurons or dropped. {Due to the stochastic nature of the NeuroSA algorithm}, these artifacts or errors can be tolerated during the initial phases of the convergence process. Asymptotically, as the network spiking rates decrease or the inter-spike interval increases, there would be enough time between events for the pseudo-gradient information to be correctly routed and hence Eq.~\ref{eq_methods_dH_async} is satisfied. This region of convergence corresponds to the low-temperature regime where it is important to explore distant states and at the same time accept proposals (or produce spikes) only when the network energy decreases. 

The tolerance of the NeuroSA architecture to communication errors provides a mechanism to accelerate its convergence using a hybrid approach as mentioned previously. In the initial phase of hybrid approach, the optimization endeavor follows the steepest gradient descent and performs parallel updates on neuron states resembling SDP. The expected value of a solution found in the initial phase can be characterized by the GW-like bounds~\cite{Goemans1995GWUGC}. Then, the search follows the brute-force approach where NeuroSA can asymptotically approach the ground state and in the process exploring novel solutions that are better than the SOTA. 
As shown in SI Fig. S5, when NeuroSA is initialized at a low temperature (less noisy threshold) initially, the architecture converges to a solution that is $>95\%$ of SOTA. The convergence in this case is $10^4$ times faster than the case when NeuroSA is initialized at a warmer temperature (more noisy threshold). To avoid getting trapped in the neighborhood of a local attractor, the system temperature is increased and annealed according to the optimal cooling schedule, as depicted in SI Fig. S5. Note that the asymptotic performance is still determined by the $\mathcal{O}(1/log)$ decay and the cold-warm acceleration does not improve the quality of the solution. Furthermore, as highlighted in Fig. 3f, as the optimization proceeds, the time needed to achieve a unit gain in the quality of the solution increases with time, with the last gain consuming the majority of the entire simulation duration. Consequently, accelerating NeuroSA's initial convergence using a low-temperature start might not significantly reduce the overall time-to-solution when the goal is to approach the asymptotic ground state. However, the approach does enhance the efficiency of the NeuroSA to approach SOTA solutions under real-time constraints. 

One of the attractive features of the NeuroSA mapping is that the architecture can be readily implemented and scaled up on existing neuromorphic platforms like SpiNNaker2, especially given the availability of large-scale systems such as the 5-million cores supercomputer in Dresden~\cite{mayr2019spinnaker2} with more than 35K SpiNNaker2 chips interconnected in a single system. The algorithmic advantage of mapping NeuroSA on a parallel architecture was demonstrated in the small-scale experiment shown in Fig. 3e. Effectively, a parallel search amplifies the chance of approaching the SOTA/ground state within finite time-interval, because of the inherent independence of the brute-force search in the low-temperature regime. This implies that multiple copies of NeuroSA can be instantiated simultaneously on SpiNNaker2 without incurring little to no overhead, as long as local memory overflow is avoided. As the problem scales, memory mapping and encoding in neuromorphic architectures like SpiNNaker2 becomes the main challenge and bottleneck. However, the inherent sparsity in the COPs (defined by the matrix $\bf{Q}$ can be leveraged for trade-off. In addition to the algorithmic advantage of NeuroSA mapping on parallel architecture, neuromorphic hardware like SpiNNaker2 also exhibits energy-to-solution and time-to-solution advantage over CPU as described in the SI Section S5.2, where the near-PE local memory and on-chip random number generator facilitates the acceleration of NeuroSA. The key bottleneck in NeuroSA and other neuromorphic architectures executing random-walk type algorithms is the process of generating the i.i.d random variables within each neuron. It has been reported that~\cite{Pierog2015pseudoEnergy}, generating high-quality random noise consumes significant energy and many neuromorphic architectures resort to physical noise (noise intrinsic in devices) as an efficient source of randomness. In our previous works~\cite{Zhou2017FNtimer,Mehta2022FNDAM,Mehta2020FNlog} we have reported a silicon-compatible device that is capable of producing $\mathcal{O}(1/\log)$ decay required by the FN annealer. The device directly implemented the equivalent circuit shown in Fig. 1e using Fowler-Nordheim tunneling barrier where the current $J$ is determined by single electrons tunneling through the barrier. Future work will investigate how to leverage these discrete single-electron events to produce the other random variables $\mathcal{N}^B_n$ and $\mathcal{N}^E_n$.    

Note that the $\mathcal{O}(1/\log)$ dynamics can either be generated using a physical FN-tunneling device or can be emulated by implementing an FN-tunneling dynamical systems model on digital hardware. In this work, we chose the digital emulation for better scalability. However, a physical FN-tunneling device could also be used as an extrinsic or intrinsic device~\cite{Zhou2017FNtimer}. The sampling and scaling of the FN-decay can be chosen arbitrarily and is only limited by the resolution of the single-electron tunneling process (quantum uncertainty). This can be mitigated by choosing a larger size device or by modulating the voltage of the device. The time scale of the $\mathcal{O}(1/\log)$ decay can therefore be adjusted to fit the scheduling presented in the algorithm. Note that even with the optimal noise schedule, higher precision sampling and annealing (threshold adjustment) would be required. In SI Fig. S4 we show the degradation in the quality of the solution when a lower precision computation is use for NeuroSA computation. Due to the requirement of higher precision sampling of the FN device, we have resorted to a digital emulation of the FN-tunneling device on the SpiNNaker2 platform. Note that our previously reported FN-devices~\cite{Zhou2017FNtimer} could have been used for this purpose, but only in conjunction with analog-to-digital converters with more than 16 bit precision. The scope of this work is to present the algorithmic advancement in developing an asynchronous neuromorphic architecture that can utilize the FN-annealing dynamics.

The NeuroSA architecture opens the possibility of using neuromorphic hardware platforms to find novel solutions by sampling previously unexplored regions of the COP landscape. Given the combinatorial nature of the problem, even a minor improvement in the quality of the solution over the SOTA solution signifies discovering a previously unknown configuration. However, our results suggest that finding such a solution requires a significant number of compute cycles or equivalently a significant expenditure of physical energy. This is evident in Fig. 3f, which plots the number of compute cycles for a unit increase in the solution metric. The trend shows an exponential/sub-exponential growth which highlights the challenge in uncovering new solutions. Because the SA algorithm itself can be slow, the benefits of NeuroSA compared to other polynomial-time COP solvers (like the GW algorithm) lie in the regime where minimal gradient information is available to guide the optimization process, and the algorithm has to resort to a brute force random search. In this region, the search becomes sub-exponential-time or exponential-time and finding the next best solution consumes all of the run-time as was shown in Fig. 3f. However, for NeuroSA, the asymptotic improvement in solution quality is guaranteed due to the FN-tunneling dynamics. This feature makes NeuroSA stand out among other COP solvers like CIM and SBM. We acknowledge that the proposed ideal neuromorphic ISING machine is a superordinate concept that can use mechanisms other than FN tunneling to achieve annealing and the desired asymptotic convergence properties. This is because there are other stochastic global optimization methods~\cite{Aimone2023GWLoihi, Aimone2022StochasticNeuronMaxcut} than just simulated annealing which can be used for the neuromorphic ISING architecture. However, when using simulated annealing, according to the results by Hajek~\cite{Hajek1988OptimalCooling} and Geman~\cite{geman1984stochastic}, the annealing schedule should follow a schedule that is slower than $c/log(t)$ (or FN tunneling) dynamics to inherit the asymptotic global optimization convergence properties. In fact, the scaling parameter $c$ needs to be selected to achieve the best trade-off between faster convergence and the quality of the solution for a given optimization problem. Also, a schedule that follows $c/\log^{1/2}(t)$ or $c/\log\log(t)$ dynamics would also guarantee asymptotic convergence to the ground state, however, at a much slower rate.
Furthermore, we would also point out that NeuroSA, like most COP solvers, is one-shot solver unlike AI/ML inference engines, where the overhead of physical instantiation can be amortised over repetitive runs. Since one of the focus of NeuroSA is to explore new solutions as it asymptotically converges to the ground state, it is expected that NeuroSA implementations would require long run-times. As new/better solutions are produced by the solver they can be read out and used in other applications.

\section{Methods}\label{sec3_methods}

\subsection{Asynchronous Ising Machine Model}\label{sec_methods_async_ising}
QUBO and Ising formulations are interchangeable through a variable transform $s \leftrightarrow \frac{1 + s}{2}$, and hence, without any loss of generality, we consider the following optimization problem 
\begin{equation}
    \min_{\mathbf{s}\in\{-1, +1\}^{D}}H(\mathbf{s})=\frac{1}{2}\mathbf{s}^\intercal\mathbf{Q}\mathbf{s}
    \label{eq_methods_ising_energy}
\end{equation}
where $\mathbf{s} = \left[s_1,s_2,..,s_D\right]$ denotes a spin vector comprising of binary optimization variables.
Because $s_j^2 = 1,\forall j=1...D$, Eq.~\ref{eq_methods_ising_energy} is equivalent to 
\begin{equation}
    \min_{\mathbf{s}\in\{-1, +1\}^{D}}H(\mathbf{s})=\frac{1}{2}\mathbf{s}^\intercal\mathbf{Q}\mathbf{s}\quad\text{with }Q_{ii} = 0
    \label{eq_methods_ising_energy_1}
\end{equation}
Note the matrix $\mathbf{Q}$ can be symmetrized by $\mathbf{Q} \leftarrow\frac{1}{2}(\mathbf{Q}+\mathbf{Q}^\intercal)$ without changing the solution to Eq.~\ref{eq_methods_ising_energy}. Let the vector $\mathbf{s}$ at time instant $n$ be denoted by $\mathbf{s}_n$ and the change in $\mathbf{s}$ be denoted as $\Delta\mathbf{s}_n$, then
\begin{equation}
    \Delta H_n = H(\mathbf{s}_{n-1}+2\Delta\mathbf{s}_n)-H(\mathbf{s}_{n-1}),
    \label{eq_methods_dH}
\end{equation}
where $\Delta\mathbf{s}_n=\{-1,0,+1\}^D$ and $\Delta s_{j,n}s_{j,n-1}=-1,0, \forall j=1...D$ ensures that the spin either flips or remains unchanged. Then,
\begin{equation}
    \Delta H_n=\Delta\mathbf{s}_n^\intercal\mathbf{Q}(\mathbf{s}_{n-1}+2\Delta\mathbf{s}_n)+\mathbf{s}_{n-1}^\intercal\mathbf{Q}\Delta\mathbf{s}_n
    \label{eq_methods_dHds}
\end{equation}
Using $Q_{ij}=Q_{ji}$,
\begin{equation}
    \Delta H_n=2\Delta\mathbf{s}_n^\intercal\mathbf{Q}(\mathbf{s}_{n-1}+\Delta\mathbf{s}_n)
    \label{eq_methods_dHds2}
\end{equation}
and applying $s_{p,n-1}\Delta s_{p,n} = -1, \forall n,p$ leads to
\begin{equation}
    \Delta H_n=2\sum_{p\in\mathcal{C}}\Delta s_{p,n}\left[\sum_{j\notin\mathcal{C}}Q_{pj}s_{j,n}\right],
    \label{eq_methods_dHds_spike}
\end{equation}
where the set $\mathcal{C}=\{i:\Delta s_{i,n}\not= 0\}$ denotes the neurons that do not fire at time-instant $n$. Solving Eq.~\ref{eq_methods_ising_energy} involves solving the sequentially sub-problem: $\forall n$, find $\Delta s_{p,n}\in\{-1,0,+1\}^D$ such that $\sum_{p\in\mathcal{C}}\Delta s_{p,n}\left[\sum_{j\notin\mathcal{C}}Q_{pj}s_{j,n}\right] < 0$, which in itself is a combinatorial problem. By adopting asynchronous firing dynamics, the problem of searching for the set of firing neurons can be simplified. For an asynchronous spiking network, only one of the neurons can emit a spike at any time instant $n$ (due to Poisson statistics), which leads to
\begin{equation}
    \Delta H_n=2\Delta s_{p,n}\left[\sum_{j=1}^DQ_{pj}s_{j,n-1}\right]
    \label{eq_methods_dH_async}
\end{equation}
where we have used $Q_{pp}=0$. Hence, $\Delta H_n < 0$, if and only if 
\begin{equation}
    \Delta s_{p,n}\left[\sum_{j=1}^DQ_{pj}s_{j,n-1}\right] < 0.
    \label{eq_methods_dH_async_criterion}
\end{equation}

\subsection{Derivation of NeuroSA's neuron model }\label{sec_methods_onoff_mapping}
In its most general form~\cite{Kirkpatrick1983}, a simulated annealing algorithm solves Eq.~\ref{eq_methods_ising_energy} by accepting or rejecting choices of $\Delta s_{p,n}$ according to 
\begin{equation}
    \text{Accept} \quad \Delta s_{p,n}: \text{if}\quad B\exp\left(\frac{-\Delta H_n}{T_n}\right)>u_n,
    \label{eq_methods_SAdynamics}
\end{equation}
where $u_n$ is a uniformly distributed r.v. between $\left[0,1\right]$, and $B>1$ is a hyper-parameter, $T_n>0$ denotes the temperature at time-instant $n$. Eq.~\ref{eq_methods_SAdynamics} is equivalent to
\begin{equation}
    \text{Accept} \quad \Delta s_{p,n}: \text{if}\quad \Delta H_n<-T_n\log\left(\frac{u_n}{B} + \epsilon\right)
    \label{eq_methods_SA_dH}
\end{equation}
or
\begin{equation}
    \Delta s_{p,n}\left[\sum_{j=1}^DQ_{pj}s_{j,n-1}\right]<-T_n \mathcal{N}_n^E
    \label{eq_methods_SA_ds}
\end{equation}
where $\mathcal{N}_n^E = \log\left(\frac{u_n}{B} + \epsilon\right)$ is an exponentially distributed r.v.. We have introduced a small additive term $\epsilon > 0$ to ensure numerical stability when drawing samples with values close to zero. In practice, $\epsilon$ is determined by the precision of the hardware platform and hence will be considered a hyperparameter for NeuroSA. Eq.~\ref{eq_methods_SA_ds} can be written case-by-case as 
\begin{equation}
    \Delta s_{p,n}=
    \left\{ 
      \begin{array}{ c l }
        +1 & \quad \textrm{if }s_{p,n-1}=-1 \textrm{  and  }-\sum_{j=1}^DQ_{pj}s_{j,n-1}>T_n\mathcal{N}_n^E\\
        -1 & \quad \textrm{if }s_{p,n-1}=+1 \textrm{  and  }\sum_{j=1}^DQ_{pj}s_{j,n-1}>T_n\mathcal{N}_n^E\\
         0 & \quad \textrm{otherwise}
      \end{array}
    \right.\label{eq_methods_firing_criteria}
\end{equation}
Decomposing the variables differentially as $\Delta s_{p,n}=\Delta s_{p,n}^+-\Delta s_{p,n}^-$, $s_{p,n}=s_{p,n}^+-s_{p,n}^-$, $\Delta s_{p,n}^+$, $\Delta s_{p,n}^-$, $s_{p,n}^+$, $ s_{p,n}^- >0$ leads to $s_{p,n}^+=\sum_{k=1}^n\left[\Delta s_{p,k}^+ - \Delta s_{p,k}^-\right]$, $s_{p,n}^-=\sum_{k=1}^n\left[-\Delta s_{p,k}^+ + \Delta s_{p,k}^-\right]$. Eq.~\ref{eq_methods_firing_criteria} is therefore equivalent to
\begin{equation}
    \Delta s_{p,n}^+=
    \left\{ 
      \begin{array}{ c l }
        1 & \quad \textrm{if }s_{p,n-1}^+=0 \textrm{  and  }\displaystyle\sum_{k=1}^{n-1}\displaystyle\sum_{j=1}^DQ_{pj}\left(-\Delta s_{j,k}^+ + \Delta s_{j,k}^-\right)>T_n\mathcal{N}_n^E\\
        0 & \quad \textrm{otherwise}
      \end{array}
    \right.\label{eq_methods_firing_criteria_on}
\end{equation}
which corresponds to the spiking criterion for an ON neuron and
\begin{equation}
    \Delta s_{p,n}^-=
    \left\{ 
      \begin{array}{ c l }
        1 & \quad \textrm{if }s_{p,n-1}^-=0 \textrm{  and  }\displaystyle\sum_{k=1}^{n-1}\displaystyle\sum_{j=1}^DQ_{pj}\left(\Delta s_{j,k}^+ - \Delta s_{j,k}^-\right)>T_n\mathcal{N}_n^E\\
        0 & \quad \textrm{otherwise}
      \end{array}
    \right.\label{eq_methods_firing_criteria_off}
\end{equation}
which corresponds to the spiking criterion for an OFF neuron.
Introducing a RESET parameter $A\gg|T_n\log\left(\frac{u_n}{B} + \epsilon\right)|$, Eq.~\ref{eq_methods_firing_criteria_on} and~\ref{eq_methods_firing_criteria_off} are equivalent to the
ON neuron model
\begin{equation}
    \Delta s_{p,n}^+=
    \left\{ 
      \begin{array}{ c l }
        1 & \quad \textrm{if }v_{p,n}^+>T_n\mathcal{N}_n^E,\\
          & \quad\textrm{where}\quad v_{p,n}^+=v_{p,n-1}^++\sum_{j=1}^DQ_{pj}(\Delta s_{j,n-1}^--\Delta s_{j,n-1}^+)\\
          & \quad\quad\quad\quad\quad+A\Delta s_{p,n-1}^--A\Delta s_{p,n-1}^+\\
        0 & \quad \textrm{otherwise}
      \end{array}
    \right.\label{eq_methods_firing_criteria_on_reset}
\end{equation}
and the OFF neuron model
\begin{equation}
    \Delta s_{p,n}^-=
    \left\{ 
      \begin{array}{ c l }
        1 & \quad \textrm{if }v_{p,n}^->T_n\mathcal{N}_n^E,\\
          & \quad\textrm{where}\quad v_{p,n}^-=v_{p,n-1}^-+\sum_{j=1}^DQ_{pj}(\Delta s_{j,n-1}^+-\Delta s_{j,n-1}^-)\\
          & \quad\quad\quad\quad\quad+A\Delta s_{p,n-1}^+-A\Delta s_{p,n-1}^-\\
        0 & \quad \textrm{otherwise}
      \end{array}
    \right.\label{eq_methods_firing_criteria_off_reset}
\end{equation}
The variables $v_{p,n}^+, v_{p,n}^-$ represent the membrane potentials of the ON-OFF integrate-and-fire neurons at the
time instant $n$.
To ensure that all neurons are equally likely to be selected (to satisfy the ergodicity property of SA), we introduce a Bernoulli r.v. for every neuron $p$ as
\begin{equation}
    \mathcal{N}_{p}^{B}=
    \left\{ 
      \begin{array}{ c l }
        1 & \quad \textrm{with probability  }1-\eta\\
        0 & \quad \textrm{with probability  }\eta
      \end{array}
    \right.\label{eq_methods_bernoulli_rv}
\end{equation}
which leads to the ON neuron model
\begin{equation}
    \Delta s_{p,n}^+=
    \left\{ 
      \begin{array}{ c l }
        1 & \quad \textrm{if }v_{p,n}^+>\mu_{p,n},\\
        0 & \quad \textrm{otherwise}
      \end{array}
    \right.\label{eq_methods_firing_criteria_on_bernoulli}
\end{equation}
and the OFF neuron model
\begin{equation}
    \Delta s_{p,n}^-=
    \left\{ 
      \begin{array}{ c l }
        1 & \quad \textrm{if }v_{p,n}^->\mu_{p,n},\\
        0 & \quad \textrm{otherwise,}
      \end{array}
    \right.\label{eq_methods_firing_criteria_off_bernoulli}
\end{equation}
where $\mu_{n,p}=T_n\mathcal{N}_n^E+A\mathcal{N}_{p,n}^B$ denotes the shared noisy threshold between the $p^{th}$ pair of ON-OFF neurons at time instance $n$. The ON-OFF construction ensures that $\Delta s_{p,n}^+ \Delta s_{p,n}^- = 0, \forall p,n$, which leads to the following fundamental ON-OFF integrate-and-fire neuron model of NeuroSA which is summarized as:
The ON-OFF neuron's membrane potentials $v_{p,n}^+, v_{p,n}^- \in \mathbb{R}$ evolve as
\begin{gather}
    v_{p,n}^+ \leftarrow v_{p,n-1}^++\sum_{j=1}^DQ_{pj}(\Delta s_{j,n-1}^--\Delta s_{j,n-1}^+) + A\Delta s_{p,n-1}^- \\
    v_{p,n}^- \leftarrow v_{p,n-1}^-+\sum_{j=1}^DQ_{pj}(\Delta s_{j,n-1}^+-\Delta s_{j,n-1}^-) + A\Delta s_{p,n-1}^+ 
    \label{eq_intro_membraneIF}
\end{gather}
where $A > 0$ is a constant that represents an excitatory synaptic coupling between the ON and the OFF neurons, as shown in Fig. 1d. The ON and OFF neurons generate a spike when their respective membrane potential exceeds a time-varying noisy threshold $\mu_{p,n}$ according to 
\begin{equation}
    \Delta s_{p,n}^+=
    \left\{ 
      \begin{array}{ c l }
        1 & \quad \textrm{if }v_{p,n}^+>T_n \mathcal{N}_{p,n}^E + A\mathcal{N}_{p,n}^B,\\
        0 & \quad \textrm{otherwise}
      \end{array}
    \right.
    \label{eq_intro_firing_criteria_on_reset}
\end{equation}
and
\begin{equation}
    \Delta s_{p,n}^-=
    \left\{ 
      \begin{array}{ c l }
        1 & \quad \textrm{if }v_{p,n}^->T_n \mathcal{N}_{p,n}^E + A\mathcal{N}_{p,n}^B,\\
        0 & \quad \textrm{otherwise}
      \end{array}
    \right.
    \label{eq_intro_firing_criteria_off_reset}
\end{equation}
after which the membrane potentials are RESET by subtraction according to
\begin{gather}
    v_{p,n}^+ \leftarrow v_{p,n}^+ - A\Delta s_{p,n}^+\\
    v_{p,n}^- \leftarrow v_{p,n}^- - A\Delta s_{p,n}^-.
    \label{eq_intro_membrane_reset}
\end{gather}
Note that the RESET by subtraction is a commonly used mechanism in spiking neural networks and neuromorphic hardware~\cite{Rueckauer2017resetSubstract}. Also, note that the asynchronous RESET of the membrane potential is instantaneous and the spike is represented by a (0,1) binary event.

\subsection{Dynamical systems model implementing the FN Annealer}\label{sec_methods_FN_annealer}

In~\cite{Hajek1988OptimalCooling} it was shown that a temperature annealing schedule of the form
\begin{equation}
    T_n\geq\frac{c}{\log\left({1+n}\right)}.
    \label{eq_methods_optimal_temp}
\end{equation}
can ensure that the simulated annealing will asymptotically converge to the ground state of the underlying COP. The parameter $c$ in Eq.~\ref{eq_methods_optimal_temp} is chosen to be larger than the depths of all the local minima of COP. The equivalent continuous-time model for the lower-bound in Eq.~\ref{eq_methods_optimal_temp} that produces $T(t)$ is given by 
\begin{equation}
    T(t)=\frac{T_0}{\log\left({1+\frac{t}{C}}\right)}
    \label{eq_methods_optimal_temp2}
\end{equation}
where $C$ is a normalizing constant. Differentiating Eq.~\ref{eq_methods_optimal_temp2} one obtains the dynamical systems model 
\begin{equation}
    C\frac{dT}{dt}= - \frac{T^2}{T_0}\exp{\frac{-T_0}{T}}
    \label{eq_methods_FN_log}
\end{equation}
that generates $T(t)$. 

The R.H.S of Eq.~\ref{eq_methods_FN_log} has the form of the current flowing across a Fowler-Nordheim quantum-mechanical tunneling junction and Eq.~\ref{eq_methods_FN_log} describes a FN integrator~\cite{Zhou2017FNtimer} with a capacitance $C$. Combining with the expressions of the dynamical threshold in Eq.~\ref{eq_methods_firing_criteria_on_bernoulli}-~\ref{eq_methods_firing_criteria_off_bernoulli}, which includes the exponentially distributed and Bernoulli distributed random variables $\mathcal{N}_{n}^E$ and $\mathcal{N}_{n}^B$, leads to the equivalent circuit model of the FN annealer shown in Fig. 1d and e. {A discrete-time equation approximating the continuous-time dynamical systems model in equation~\ref{eq_methods_FN_log} was implemented on the CPU or the SpiNNaker2 platform. The sampling period in the discrete-time model was chosen by adjusting $C$ and to fit the optimal scheduling requirements of the NeuroSA algorithm.} For all experiments, $C=8\times10^4$, and $T_0=0.3125$. The mean of the random variable $\mathcal{N}_{p,n}^E$ was chosen to be $-0.916$.  

\subsection{Acceleration of NeuroSA on Synchronous and Clocked Systems}\label{sec_methods_sync_impl}
While the ideal implementation of NeuroSA architecture requires a fully asynchronous architecture, most neuromorphic accelerators are either fully clocked and use address-event-routing-based packet switching, such as HiAER-spike, or employ globally asynchronous interrupt-driven units that are locally clocked (synchronous), such as SpiNNaker2. For
these clocked systems, NeuroSA can be efficiently implemented by exploiting the mutual independence and i.i.d. properties of the r.vs $\mathcal{N}^E$
and $\mathcal{N}^B$. The SI Section S4 describes the pseudo-code for CPU and SpiNNaker2 implementations. In the NeuroSA architecture, each neuron determines its spiking behavior solely from its internal parameters, i.e. the membrane potential, neuron state, etc. Therefore, $\mathcal{N}^E_{p,n}$ needs to be distinct and local to each neuron in the system. On the other hand, the ergodicity of the optimization process can be enforced through a global arbiter. We decouple the Bernoulli noise from the noisy threshold such that only the decision threshold $\mu^*_{p,n} = T_n\mathcal{N}_{p,n}^E$ is applied to each neuron. In this case, multiple spikes may occur at each simulation step. All the neurons that emit spikes synchronously are referred to as active neurons. Out of all active neurons, only one gets selected by the global arbiter and propagated to other neurons, while the remaining spikes are discarded. This inhibitive firing dynamics ensures that at most one spike is transmitted and processed, which satisfies the asynchronous firing requirement as shown in Eq.~\ref{eq_methods_dH_async}. The global arbiter is implemented differently across the neuromorphic hardware that we have tested on, with the detailed implementation documented in SI Section S4 for CPU (software) implementation, and Section S5 for SpiNNaker2 implementation. 

\subsection{Generation of Network PCA Trajectories} \label{sec_methods_PCA}
To demonstrate and visualize the evolution of the network dynamics for a large problem, we used Principal component analysis (PCA) to perform dimensionality reduction on the population dynamics similar to a procedure reported in literature~\cite{friedrich2001dynamic, stopfer2003intensity, gangopadhyay2021gtnn}. In NeuroSA, the population spiking activity indicates changes in the neuronal states, the attractor dynamics in proximity to a local/global minimum, and the escape mechanisms for exploring the state space. As shown in Fig. 1h, the spikes across the neuronal ensembles are binned within a predefined time window to produce a real-valued vector. The time window is then shifted with some pre-defined overlap to produce a sequence of real-valued vectors. PCA is then performed over all the real-valued vectors and only the principal vectors with largest eigenvalues are chosen. The real-valued vector sequence is then projected onto the three principal components resulting in 3D trajectories shown in Fig. 1i and 2d.

{\subsection{Estimation of empirical probability density function} \label{sec_methods_pdf}}
{For the MAX-CUT and MIS experiments, NeuroSA was run on each graph for 5 times. The results in each run were normalized with respect to the SOTA and then grouped as a histogram. The probability density function (pdf) was generated using ksdensity, the built-in MATLAB function. }

\subsection{Data Availability}\label{sec_data}
The COP solutions, the algorithm execution time, the power and the energy data for the CPU and SpiNNaker2 imeplementation of NeuroSA generated in this study are available at https://github.com/aimlab-wustl/neuroSA.

\subsection{Code Availability}\label{sec_code}
The specific MATLAB and Python codes used in simulation/emulation studies on the CPU platform are available at https://github.com/aimlab-wustl/neuroSA. Since the SpiNNaker2 stack is still under development, and the system is proprietary, the software implementation on SpiNNaker2 will not be made available publicly but will be available upon requests.


\section{Acknowledgements}\label{sec_acknowledgements}
This work is supported in part by research grants from the US National Science Foundation: ECCS:2332166 and FET:2208770. M.A., H.G., G.L., S.M. and C.M. would like to acknowledge the financial support by the Federal Ministry of Education and Research of Germany in the programme of “Souver\"an. Digital. Vernetzt.”. Joint project 6G-life, project identification number: 16KISK001K. M.A., H.G, G.L., S.M. and C.M. acknowledge the EIC Transition under the ”SpiNNode” project (grant number 101112987). M.A., H.G, G.L., S.M. and C.M. also acknowledge the Horizon Europe project "PRIMI" (Grant number 101120727). G.U. acknowledges a contribution from the Italian National Recovery and Resilience Plan (NRRP), M4C2, funded by the European Union –NextGenerationEU (Project IR0000011, CUP B51E22000150006, “EBRAINS-Italy”).

\section{Author Contributions}\label{sec_contributions}
Z.C., Z.X., M.A., J.L., O.O., A.M., N.D., C.H., S.B., J.E., R.P., G.U., A.G.A., S.S., C.M., G.C., and S.C. participated in a workgroup titled Quantum-inspired Neuromorphic Systems at the Telluride Neuromorphic and Cognitive Engineering (TNCE) workshop in 2023, and the outcomes from the workgroup have served as the motivation for this work. S.C. formulated the asynchronous ON-OFF neuron model with the FN-annealer; Z.C. and S.C. designed the NeuroSA experiments; Z.C. benchmarked NeuroSA on different MAX-CUT graphs; Z.X. implemented the first version of SA algorithm;  S.C, Z.C., J.L and G.C proposed the use of spike events to reduce communication bottleneck in NeuroSA; M.A., H.G, C.M., S.M., G.L., and Z.C implemented NeuroSA on SpiNNaker2; M.A., S.M., and G.L. optimized SpiNNaker2 for MAX-CUT benchmarks; All authors/co-authors contributed to proof-reading and writing of the manuscript.  

\section{Competing interests}\label{sec_COI}
SpiNNaker2 is a neuromorphic hardware accelerator platform by SpiNNcloud Systems, a commercial entity with whom M.A., H.G., S.M., G.L., and C.M. have affiliations and financial interests. S.C. is named as an inventor on U.S. and international patents associated with FN-based dynamical systems, and the rights to the intellectual property are managed by Washington University in St. Louis. The remaining authors declare no competing interests.

\newpage
\setcounter{page}{1}
\setcounter{section}{0}
\renewcommand{\thesection}{S\arabic{section}}

\makeatletter
\renewcommand\theHsection{supp.\thesection}
\makeatother
\setcounter{section}{0}
\renewcommand{\thesection}{S\arabic{section}}
\setcounter{figure}{0}
\renewcommand{\thefigure}{S\arabic{figure}}
\setcounter{table}{0}
\renewcommand{\thetable}{S\arabic{table}}

\makeatletter
\renewcommand\theHsection{supp.\thesection}
\renewcommand\theHfigure{supp.\thefigure}
\makeatother
\resetlinenumber[1]
\title{ON-OFF Neuromorphic ISING Machines using Fowler-Nordheim Annealers}
\section*{Supplementary Information}\label{sec_supp}

\section{General Ising Model with $\mathbf{b} \ne 0$}\label{sec_supp_1}

Given a spin vector $\mathbf{s} = \left[s_1,s_2,..,s_D\right]$ and an external bias vector (or a field) $\mathbf{b}$, a general Ising Hamiltonian has the form: 
\begin{equation}
    \min_{\mathbf{s}\in\{-1, +1\}^{D}}H(\mathbf{s})=\frac{1}{2}\mathbf{s}^\intercal\mathbf{Q}\mathbf{s} + \mathbf{b}^T\mathbf{s}
    \label{eq_methods_ising_energy_supp}
\end{equation}
Then, following the steps in the Methods section~\ref{sec3_methods} leads to 
\begin{equation}
    {\Delta H_n=2\sum_{p\in\mathcal{C}}\Delta s_{p,n}\left[\left(\sum_{j=1}^{D}Q_{pj}s_{j,n} \right)+ b_p\right],}
    \label{eq_methods_dHds_spike_supp}
\end{equation}
where the set $\mathcal{C}=\{i:\Delta s_{i,n}\not= 0\}$. Introducing a non-spiking static neuron whose state $s_0 = 1$ remains constant, then Eq.~\ref{eq_methods_dHds_spike_supp} can be written as
\begin{equation}
    \Delta H_n=2\sum_{p\in\mathcal{C}}\Delta s_{p,n}\sum_{j=0}^{D}Q_{pj}s_{j,n},
    \label{eq_methods_dHds_spike2_supp}
\end{equation}
which has the same form as Eq.~\ref{eq_methods_dHds_spike} but with $Q_{p,0} = b_p$. {Alternatively, the external field $b_p$ for neuron $p$ can be interpreted as a shift in the membrane potential which can be programmed when initializing the neurons.}


\section{Mapping of MAX-CUT to Ising Model}\label{sec_supp_2}
{Consider a generic MAX-CUT problem on graph, $G=\{V, \mathbf{W}\}$, where $V=\{v_i|i\in1...D\}$ denotes the set of $D$ vertices and $\mathbf{W}\in\{-1,0,1\}^{D\times D}$ denote the weighted adjacency matrix. Each of the vertices $v_i$ is connected to any other vertex $v_j$ through connection $W_{ij}$. A cut on $G$ partitions the set of vertices $V$ into sets $L$ and $R$. The vertices belong to different sets can be described by an additional variable $s_i$ associated with each node $v_i$, following}
\begin{equation}
    s_i=
    \left\{ 
      \begin{array}{ c l }
        +1 & \quad v_i\in L\\
        -1 & \quad v_i\in R.
      \end{array}
    \right.\label{eq_supp_ising_var}
\end{equation}
The edge weight $W_{ij}$ connecting $v_i$ and $v_j$ is cut only when $s_is_j = -1$. MAX-CUT problem aims to maximize the number of cuts which is given by
\begin{equation}
    \max_{s_i,s_j\in\{-1,+1\}}H_{max}=\frac{1}{2}\sum_{i,j}^DW_{ij}(1-s_is_j).
    \label{eq_supp_mc_maxH}
\end{equation}
{Maximizing the total number of cuts is therefore equivalent to minimizing the following Ising energy function with the Hamiltonian matrix $\mathbf{Q}$ equaling to the weighted adjacency matrix $\mathbf{W}$,} 
\begin{equation}
    \max_{\mathbf{s}\in\{-1,+1\}^D}H_{Ising}=\frac{1}{2}\mathbf{s}^\intercal \mathbf{Q}\mathbf{s}.
    \label{eq_supp_mc_minH}
\end{equation}

\section{{Mapping of MAX-INDEPENDENT SET to Ising Model}}\label{sec_supp_3}

{The Max-Independent Set (MIS) problem on a graph \( G = \{V, \mathbf{W}\} \) can also be mapped to an Ising model. In this context, \( V = \{v_i | i \in 1 \ldots D\} \) represents the set of \( D \) vertices, and \( \mathbf{W} \in \{0, 1\}^{D \times D} \) denotes the unweighted adjacency matrix, where \( W_{ij} = 1 \) if there is an edge between vertex \( v_i \) and vertex \( v_j \), and \( W_{ij} = 0 \) otherwise. In the MIS problem, the goal is to find the largest subset of vertices \( T \subseteq V \) such that no two vertices in \( T \) are adjacent in $\mathbf{W}$. An additional variable $x_i$ can be associated to any node $v_i$, such that it represents if the node is in subset $T$ as}
\begin{equation}
{
    x_i =
    \left\{
    \begin{array}{cl}
        +1 & \quad v_i \in T \\
        0 & \quad v_i \notin T.
    \end{array}
    \right.
    }
    \label{eq_supp_ising_mis_var}
\end{equation}
{while the independent condition is satisfied such that $\forall \, v_i, v_j \in T, \: W_{ij} = 0$.} 

{Therefore to enforce the aforementioned independent constraint and maximizing the size of the independent set, the objective function can be formulated as}
\begin{equation}
    {\max_{x_i,x_j\in\{0,+1\}}H_{mis}=\left(\sum_{i}^Dx_ix_i\right)-\beta\left(\sum_{i,j}^D W_{ij}x_ix_j\right),}
    \label{eq_supp_mis_maxH}
\end{equation}
{where $\beta\in(0, 1)$ is a hyperparameter for applying penalty on solutions that violate the constraint. The Ising Hamiltonian $\mathbf{Q}$ can be formulated as $Q_{ij}=W_{ij},\:Q_{ii}=-1, \:\forall i,j\in {1...D}$. Therefore, maximizing the independent set is equivalent to minimizing the Ising energy in Eq.~\ref{eq_supp_mc_minH}, where the variable $\mathbf{x}$ can be mapped to spin states $\mathbf{s}$ following $\mathbf{s}=2\mathbf{x}-1$. Different from the formulation presented in the Methods Section~\ref{sec_methods_async_ising}, since $Q_{ii}\ne0$ in the MIS formulation, the pseudo-gradient changes as the following,
\begin{equation}
    {\Delta H_n=2\left(\Delta s_{p,n}\left[\sum_{j=1}^DQ_{pj}s_{j,n-1}\right]+Q_{pp}\right),}
    \label{eq_supp_dH_async}
\end{equation}
where we have used {$\Delta s_{p,n}^2=1$}. Hence, $\Delta H_n < 0$, if and only if 
\begin{equation}
    {\Delta s_{p,n}\left[\sum_{j=1}^DQ_{pj}s_{j,n-1}\right]+Q_{pp} < 0.}
    \label{eq_supp_dH_async_criterion}
\end{equation}}

\section{Algorithmic Implementation of NeuroSA}\label{sec_supp_4}
The NeuroSA architecture is simulated on a CPU platform using the MATLAB R2022a software package. The ON-OFF neuron pair parameters, $\mathbf{v}^\pm$, $\mathbf{s}^\pm$, and $\Delta\mathbf{s}^\pm$ are stored in pre-allocated arrays. As described in Methods section~\ref{sec_methods_sync_impl}, the Bernoulli random variable $\mathcal{N}^B_n$ can be decoupled from the firing threshold $\mu_n$ such that the threshold for each neuron pair only emulates the simulated annealing acceptance/rejection dynamics while the ergodicity is enforced by using a global random selection arbiter. The software simulation follows this implementation by generating an array of i.i.d random numbers and calculating the decision threshold accordingly. The pseudo-code for the NeuroSA software is presented as the following
\begin{algorithm}
\caption{NeuroSA Pseudo-code}\label{algo_supp_neuroSA_software}
\begin{algorithmic}
\State $\mathbf{s}^+ \gets 1$, $\mathbf{s}^- \gets 0$
\Comment{Spin states initialization}
\State $\Delta\mathbf{s}^+ \gets 0$, $\Delta\mathbf{s}^- \gets 0$
\Comment{Spikes initialization}
\State $\mathbf{v}^+ \gets \mathbf{Q}(\mathbf{s}^+-\mathbf{s}^-)$, $\mathbf{v}^- \gets -\mathbf{Q}(\mathbf{s}^+-\mathbf{s}^-)$
\Comment{Membrane potential initialization}
\State$iter\gets1$
\State$t\gets1$
\While{$iter < \text{MAX\_ITER}$}
\State $\mathbf{thld}\gets d*\left(\frac{\beta\log{U(0,1)}}{1+\log{(-\alpha t)}}\right)$
\Comment{Distinct threshold for each neuron pair}
\For{$i=1$ to $D$}
    \If{$\mathbf{s}^+[i]=0$ and $\mathbf{v}^+[i]>\mathbf{thld}[i]$}
    \Comment{ON neuron firing criteria}
        \State $\Delta\mathbf{s}^+[i]\gets1$
        \State $\Delta\mathbf{s}^-[i]\gets0$
    \ElsIf{$\mathbf{s}^-[i]=0$ and $\mathbf{v}^-[i]>\mathbf{thld}[i]$}
    \Comment{OFF neuron firing criteria}
        \State $\Delta\mathbf{s}^+[i]\gets0$
        \State $\Delta\mathbf{s}^-[i]\gets1$
    \Else
        \State $\Delta\mathbf{s}^+[i]\gets0$
        \State $\Delta\mathbf{s}^-[i]\gets0$
    \EndIf
\EndFor
\State randomly select neuron $p$ from $(\Delta\mathbf{s}^+-\Delta\mathbf{s}^-)\ne0$
\Comment{Inhibitive firing}
\State $\mathbf{s}^+[p]\gets\mathbf{s}^+[p]+\Delta\mathbf{s}^+[p]-\Delta\mathbf{s}^-[p]$
\State $\mathbf{s}^-[p]\gets\mathbf{s}^-[p]-\Delta\mathbf{s}^+[p]+\Delta\mathbf{s}^-[p]$
\State $\mathbf{v}^+ \gets \mathbf{v}^+ + 2\mathbf{Q}[p,:]\Delta\mathbf{s}^+-2\mathbf{Q}[p,:]\Delta\mathbf{s}^-$
\State $\mathbf{v}^- \gets \mathbf{v}^- - 2\mathbf{Q}[p,:]\Delta\mathbf{s}^++2\mathbf{Q}[p,:]\Delta\mathbf{s}^-$
\State $t\gets t+dt$
\EndWhile
\end{algorithmic}
\end{algorithm}

Here, MAX\_ITER denotes the maximum simulation time in discrete steps, $dt$ is the granularity of the time step, and $\alpha, \beta$ are the hardware-related hyperparameter of the FN annealer as discussed in Methods section~\ref{sec_methods_FN_annealer}. This implementation faithfully recovers the asynchronous NeuroSA architecture in that it instantiates distinctive noisy thresholds for every individual neuron pair. However, for large-scale implementation, the simulation runtime is determined by the random generation function~\cite{Pierog2015pseudoEnergy}. Therefore, we implemented a more lightweight software for large-scale simulation that reduces the footprint for random number generation. The neurons are randomly marked for selection, regardless of their firing status (active neurons). Then based on the threshold $\mu_n$, the selected neuron could fire based on the spiking criterion. This implementation is different from the synchronous implementation Algorithm~\ref{algo_supp_neuroSA_software} because only one noisy threshold is generated at a given time-instant, which reduces the CPU runtime. 

\section{Mapping NeuroSA on SpiNNaker2}\label{sec_supp_5}

\begin{figure}[h]
\centering
\includegraphics[width=\textwidth]{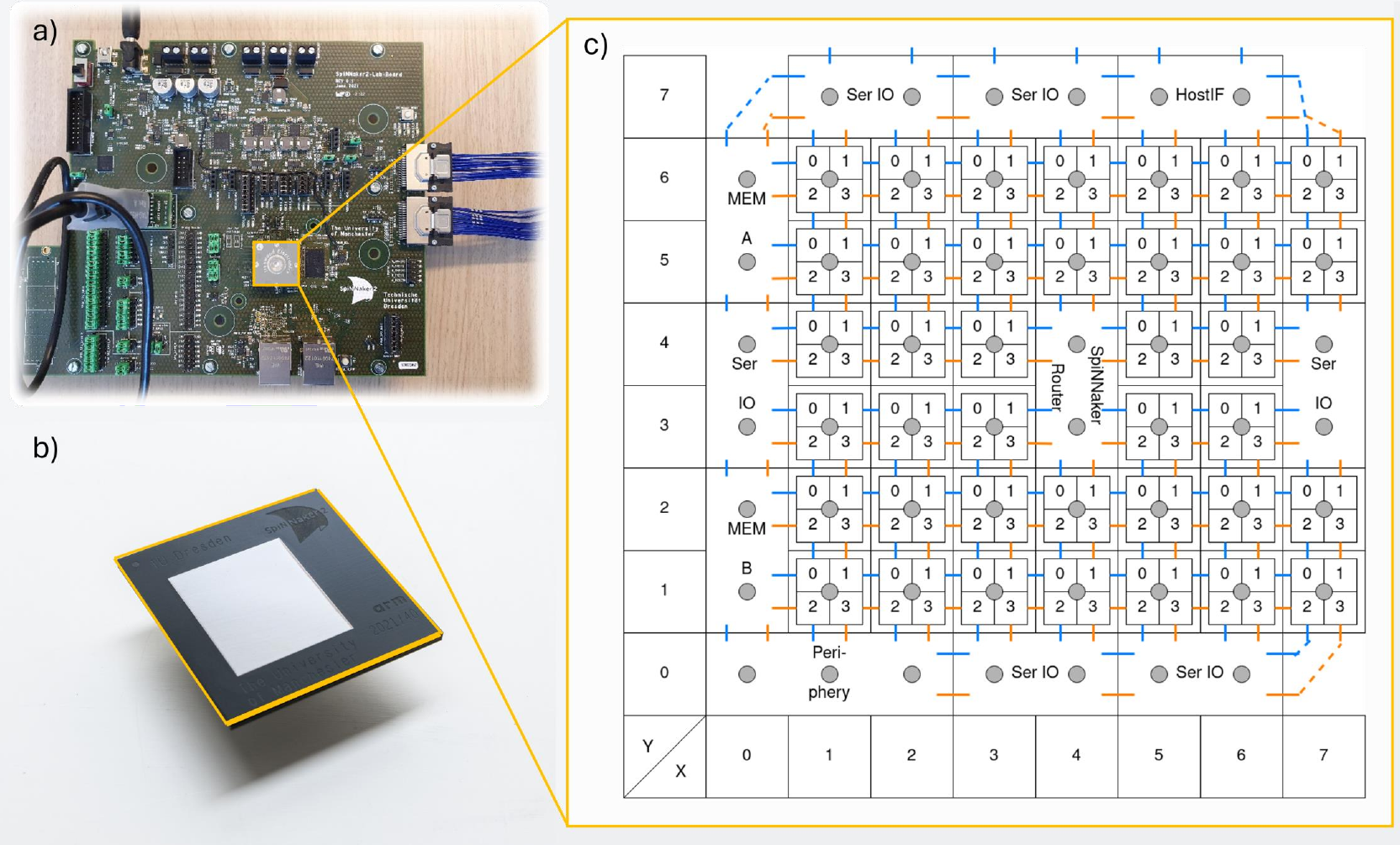}
\caption{(a) Testing board used for the NeuroSA experiments, (b) highlighting its single SpiNNaker2 chip (c) and its internal topology.}\label{nc_figspinnaker2}
\end{figure}

SpiNNaker2 is a MultiProcessor System on Chip (MPSoC) in 22nm FDSOI technology designed to execute event-based machine learning, neuromorphic, and hybrid models \cite{gonzalez2024spinnaker2}. SpiNNaker2-based systems are intended to be scaled up from one standalone chip composed by 152 Arm-based Processing Elements (PEs), to supercomputer levels with millions of PEs interconnected in a Globally Asynchronous Locally Synchronous configuration. A single chip (Fig. \ref{nc_figspinnaker2}b) contains 38 Quad-core Processing Elements, each of which employs four Arm-based PEs with custom accelerators. All the resources within a single chip are interconnected via a light-weight Network-on-Chip (NoC) as in Fig. \ref{nc_figspinnaker2}c, and operate under an interrupt-driven approach for dynamically managing power consumption. As a neuromorphic system, SpiNNaker2 implements neuron models via precompiled software that is executed in the Arm-based PEs, and uses its native communication infrastructure to redirect the spike-based activity across the system. A SpiNNaker router within each chip is the responsible component to extend the interdependent hierarchies to multi-chip levels, and beyond that to multi-board, multi-frames, and multi-rack levels. The overall topology of SpiNNaker2-based systems employs a toroidal mesh with each chip connecting to six neighbors via a predefined configuration that ensures the short communication delays within the system. SpiNNaker2 is among the most flexible neuromorphic chips providing customizations in both communication and computation to deploy more than 10 billion neurons (i.e., more than 1,000 per PE) and beyond 10,000 synapses (i.e., more than 1 million per core) in a single system~\cite{mayr2019spinnaker2}. 

\begin{figure}[t]
    \centering
    \includegraphics[width=\textwidth]{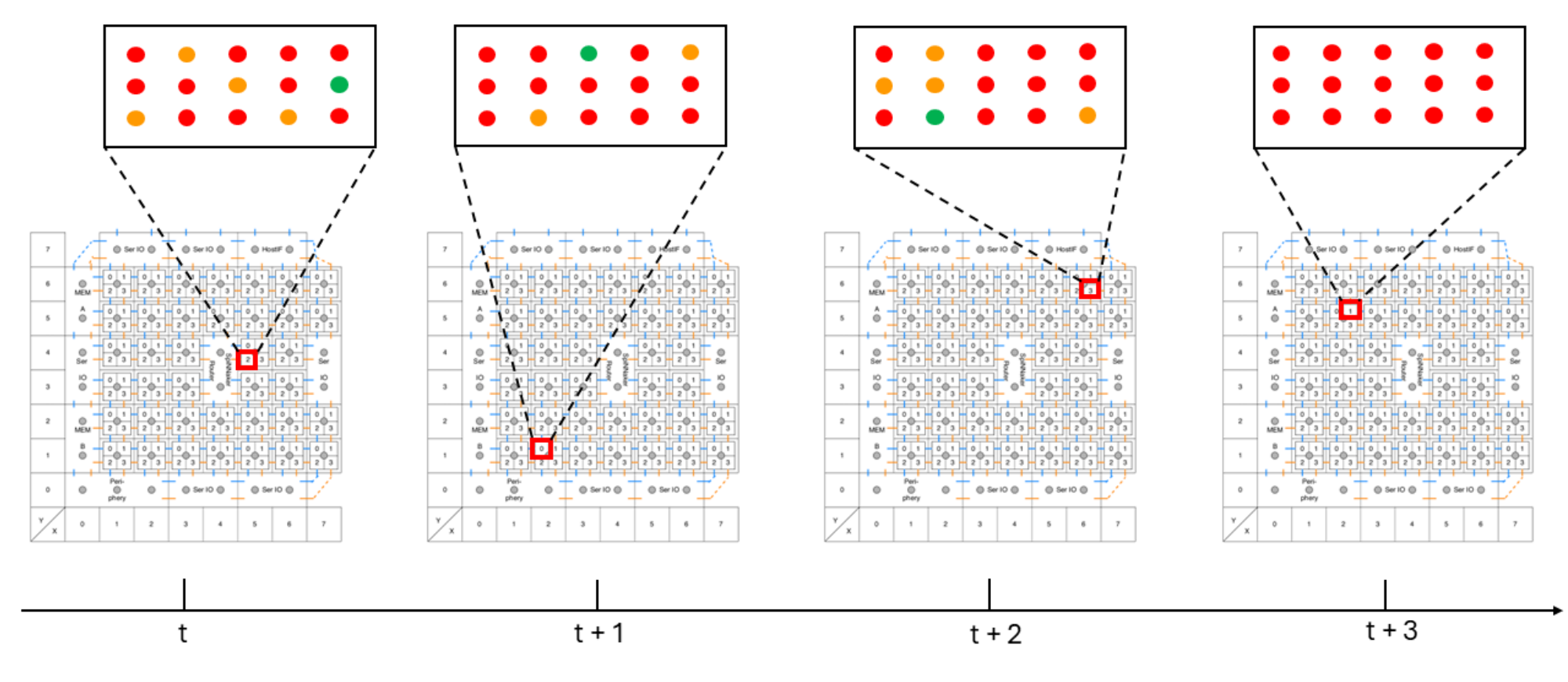}
    \caption{Overview of the random sampling of cores and neurons on SpiNNaker2. Each time step displays the internal topology of SpiNNaker2, with red squares indicating the selected core emitting a spike at that time step. Zooming in on the chosen reveals its neurons: those in red have not crossed the threshold, while those in yellow have. The green neuron signifies the spike emitter. At the final time step (t+3), no neurons have crossed the threshold, resulting in no spike emission.}
    \label{fig:spinnaker_random_sampling}
\end{figure}

The neuron model in NeuroSA is implemented through embedded software on SpiNNaker2 cores, while the high-level control and experiment configuration are performed through py-spinnaker2 \cite{vogginger_2023_10202110}, a Python library and high-level API designed for programming SpiNNaker2. Given an input MAX-CUT graph, we construct a NeuroSA network, where a mapper module within py-spinnaker2 determines the number of cores used as well as the distribution of neurons per core. This mapping depends on the network’s size, as measured by the number of neurons and synaptic connections. Following the architecture of the synchronous NeuroSA, as outlined in Methods section~\ref{sec_methods_sync_impl}, we designed a global arbiter to perform the outer-loop level random selection across all active neurons at any time step. The arbiter uniformly samples one core ID from all used core IDs per time step. This selected core is the only core that is allowed to emit a spike at that time step. Following core selection, we update the membrane potentials for all neurons. Among neurons whose membrane potentials crossed the threshold on the selected core, the global arbiter uniformly samples only one neuron to spike. If none of the neurons on the selected core crossed the threshold, no spike is emitted at that time step. The uniform sampling of the cores as well as neurons that are selected for spike emission is done using the on-chip true random number generator \cite{Neumarker2016RNG}. Fig. ~\ref{fig:spinnaker_random_sampling} depicts the firing dynamics of the NeuroSA implementation on SpiNNaker2.

\subsection{{Bare metal Implementation on SpiNNaker2}}\label{sec_supp_5_1}
{
We developed a bare metal NeuroSA implementation on the ARM micro-processors that bypassed the SpiNNaker2 software stack. As a result, the run time is significantly optimized comparing to the SpiNNaker2 software implementation($\sim40$$\times$).} 

{The NeuroSA neuron is compiled directly into micro-code that gets executed by the ARM PE cores, without going through the hierarchical general-purpose software stack. Given an input graph, neurons (nodes) are allocated to cores while ensuring that their memory requirements remain within the 128 KB limit of the local PE SRAM, taking the graph's density into account. For denser graphs, fewer neurons can be stored on a single PE due to the increased size of the routing table required to store fan-out neuron indices. The mapping is done in a static fashion where little load balancing is performed across cores. Since the communication overhead within cores are negligible and the inter-core communication creates bottleneck for transmission delay, the static mapping may account for the delay in execution in the current implementation.}

{Alongside the neuron cores, one core is designated as an orchestrator. The orchestrator PE samples a random neuron at each time step using the on-chip TRNG. It then sends a core-to-core SpiNNaker packet to the corresponding core, including the selected neuron’s index in the payload, to trigger computation.}
{Each of the neurons instantiated on a PE stores the fan-in address table to the RAM that is local to the core. This design simplifies the inter-core communication protocol: the asynchronous spikes are transmitted to the post-synaptic neurons sequentially the receiving neurons are responsible for checking and re-weighting the incoming spike and update their membrane potentials respectively. However, the sequential transmission spikes in the current implementation undermines the in-built parallelism on neuron state updates provided by the NeuroSA architecture. The lack of multi-casting ability may also be one of the reasons for the execution time of the current implementation.}

{Ongoing and future efforts will focus on identifying optimal strategies for mapping neurons onto cores, tailored to the size and density of the input graph, to minimize execution time by balancing the computation-communication tradeoff. Furthermore, this work could be enhanced by leveraging the SpiNNaker 2 communication infrastructure, such as utilizing multicast packets to broadcast information to multiple cores simultaneously.}

\subsection{Neuromorphic Advantage on SpiNNaker2}\label{sec_supp_5_2}
\begin{figure}[h]
    \centering
    \includegraphics[width=0.9\textwidth]{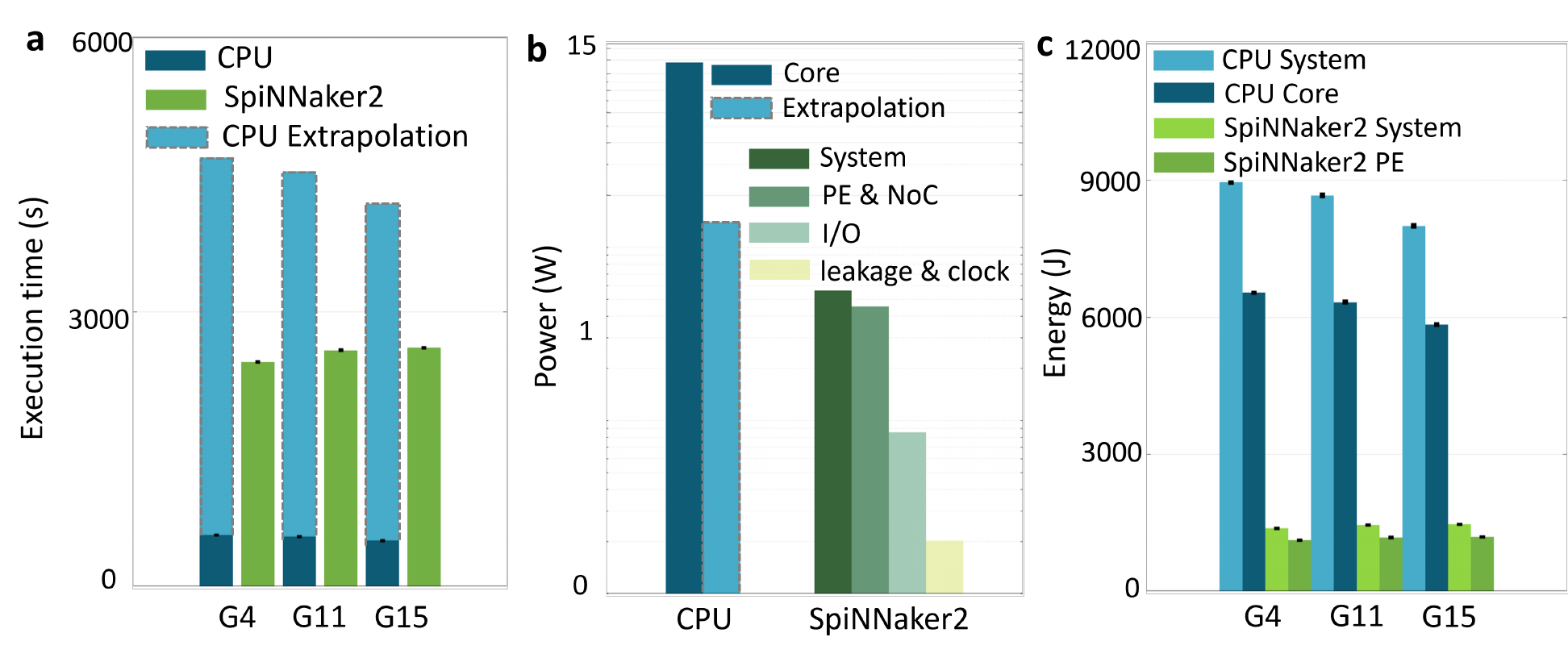}
    \caption{Presents the timing and energy metrics on both CPU and SpiNNaker2 systems that executes the same-sized NeuroSA MAX-CUT workloads (800 nodes). (a) Plots the real time-to-solution in seconds for both systems and extrapolated CPU time with respect to SpiNNaker2 clock frequency. (b) Reports the instantaneous power consumption and breakdown of it for SpiNNaker2 and CPU system. (c) Plots the energy-to-solution for both platforms, where the error bars show the maximum and minimum energy dissipation across runs.}
    \label{fig_supp_t2s}
\end{figure}
{We compare the time-to-solution for both CPU and SpiNNaker2 platforms when NeuroSA is run for the same number of execution cycles. The result is shown in Fig.~\ref{fig_supp_t2s}\textbf{a}. The CPU platform comprises of an Intel\textsuperscript{\textregistered} Core Ultra 9 185H processor core running at a maximum clock rate of $2.5$ GHz whereas the SpiNNaker2 ARM core running at a clock rate of $300$MHz. The number of iterations for NeuroSA execution is fixed across multiple runs on both the platforms and for different graphs. 
In addition to reporting the the time-to-solution for both platforms, to ensure fair comparison, we linearly extrapolated the CPU time-to-solution to account for the difference between the CPU and SpiNNaker2 clock frequencies. The result is shown in Fig.~\ref{fig_supp_t2s}\textbf{a}. The extrapolation is also justified because the NeuroSA algorithm searches the optimization landscape sequentially as the annealing schedule changes over time. The linear extrapolation serves as an optimistic lower bound on the CPU run time with lower clock frequency since it ignores the additional cascading delay that arises in the CPU pipeline due to the combination of same-speed memory and slower-clock CPU. Although the current implementation on SpiNNaker2 surpasses the CPU in terms of absolute run time due to network communication bottleneck and the sequential transmission of spiking events, the extrapolated CPU run time at equal clock frequency is worse than the SpiNNaker2 execution time.

Furthermore, we measured the NeuroSA power consumption on both CPU and SpiNNaker2 when solving the same benchmark COP. For the CPU system, we report the system and the CPU core power consumption separately. We used HWiNFO\textsuperscript{\textregistered} and Intel SOC Watch along with the Vtune Profiler~\cite{Pierro2024LoihiIsing} for the power measurement. While executing the NeuroSA algorithm, the CPU system consumes $15.93$W, in which the CPU core takes up $11.63$W. We perform similar linear extrapolation on the CPU core power consumption with respect to the SpiNNaker2 clock frequency. The extrapolation on CPU core power consumption indicate that such a core running at 300MHz would consume $1.39$W. On the other hand, SpiNNaker2 NeuroSA implementation consumes $561.9$mW. The built-in hardware sensors measures the breakdown of the power consumption as the following: the PEs (Processing Elements) and the Network on Chip (NoC) consumes $455.95$mW, the I/O $85.7$mW, and the rest $20.26$mW that accounts for unused DRAM leakage and clock generation. The power analysis is shown in Fig.~\ref{fig_supp_t2s}\textbf{b}, where SpiNNaker2 demonstrate clear power advantage even to the extrapolated results.

Next, we show the energy-to-solution results for both platforms without performing extrapolation. As shown in Fig.~\ref{fig_supp_t2s}\textbf{c}, for three different 800-node MAX-CUT problem, we plot the energy-to-finish for both platforms on the scale of entire system and comupting elements. As shown in the figure, for either the system or the processing unit, SpiNNaker2 triumphs CPU in terms of energy for completing the same NeuroSA workloads. 
Despite some non-optimality in the current SpiNNaker2 implementation of NeuroSA, as described in Supplementary Section~\ref{sec_supp_5_1}, power-efficient neuromorphic hardware already exhibits certain energy benefits in executing the NeuroSA algorithm.

\section{Effect of finite precision on NeuroSA}\label{sec_supp_precision}
The evolution of the threshold $\mu$ is the key to the robustness of the NeuroSA architecture. While CPU (or software) implementation can use floating-point precision, in practice many neuromorphic hardware accelerators support only finite precision arithmetic. 
The long-term vision is to deploy NeuroSA on custom-ASIC or a hardware platform to achieve high energy-efficiency and low time-to-solution. One option is to implement the neuron and network model using standard neuromorphic architectures such as Loihi or SpiNNaker2, where as the FN annealers are realized in analog or using mixed-signal approaches using an analog-to-digital or digital-to-analog converters.  
Here we explore how the NeuroSA performance degrades when the precision of the computation is reduced. The state of the neurons are all binary variables taking values in $\{-1, +1\}$, where the weights (or connectivity) are also quantized. Therefore, the integrate-and-fire dynamics produces membrane potentials that are also discrete integers, however, their range and precision are limited by the network size and fan-out. Therefore, the only component in the architecture that is affected by quantization is the firing threshold for each neuron. We applied quantization to the thresholding function before determining the spiking activity of a particular neuron and plotted the obtained solution for each precision. 
\begin{figure}[h]
\centering
\includegraphics[width=0.95\textwidth]{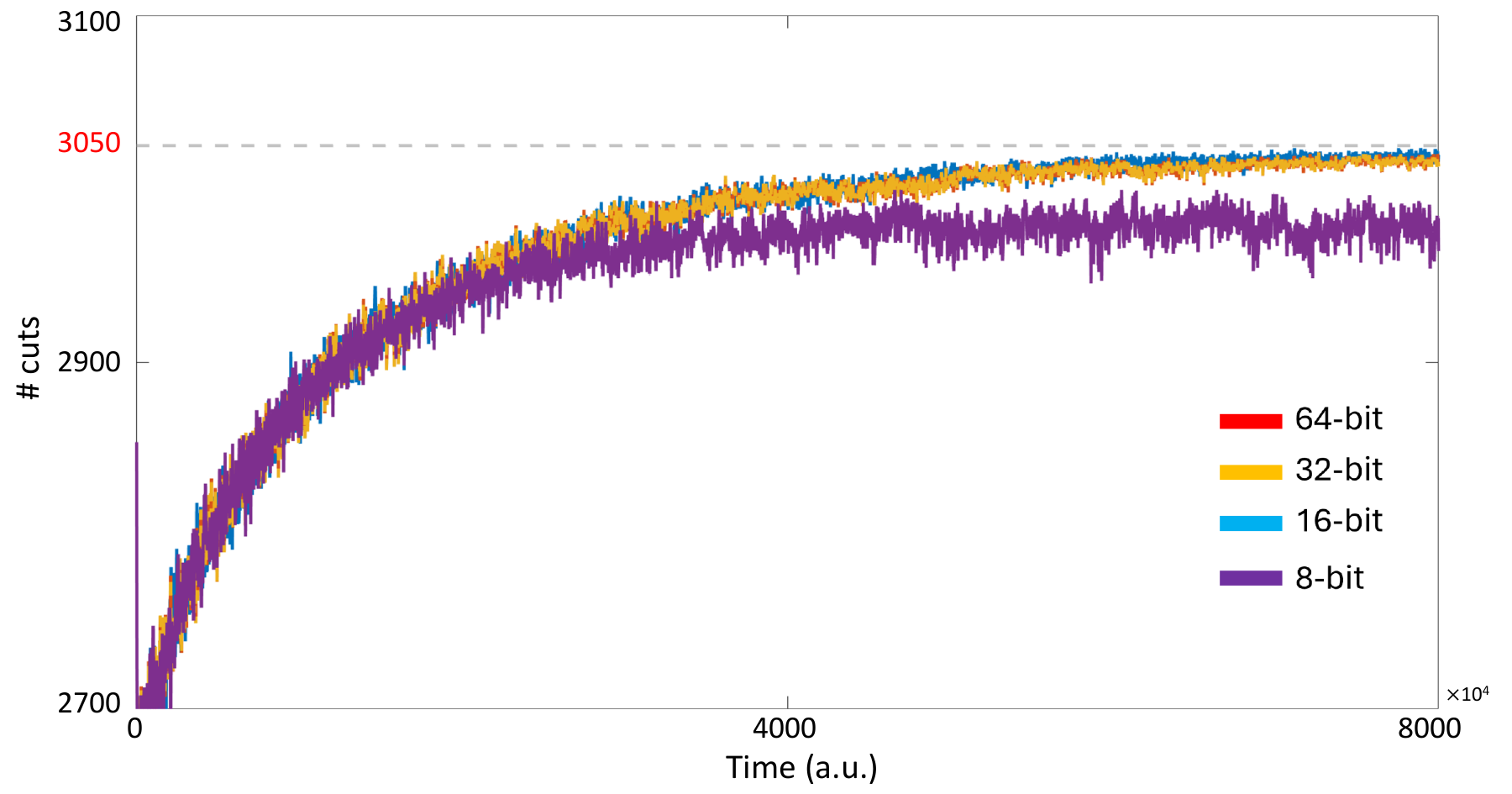}
\caption{Effect of finite precision on the performance of NeuroSA. }\label{fig_supp_precision}
\end{figure}

As shown in Fig.~\ref{fig_supp_precision}, when quantized to 64- and 32-bit floating point, the noisy thresholds incurs identical neuron population dynamics, resulting in exactly the same evolution of the obtained results. When the precision is decreased to 16-bit the network dynamics differ from the high-precision cases but the overall performance in terms of solving the original optimization problem is similar, approaching the SOTA solution. However, when the precision is further decreased to 8-bit, the performance drops at the low-temperature region, as shown in Fig.~\ref{fig_supp_precision} the purple curve. This result is as what we expected since the random fluctuation on the noisy threshold is vital to the performance of the NeuroSA architecture. When the temperature cools down, the amplitude of this fluctuation is also annealed. Under a low-precision scenario, the effect of the firing threshold fluctuation is concealed by the quantization effect. Therefore, the overall NeuroSA dynamics fail to follow the optimal simulated annealing dynamics which results in worse performance than the higher precision implementation.

\section{NeuroSA Low-temperature Start}\label{sec_supp_7}
\begin{figure}[h]
\centering
\includegraphics[width=0.95\textwidth]{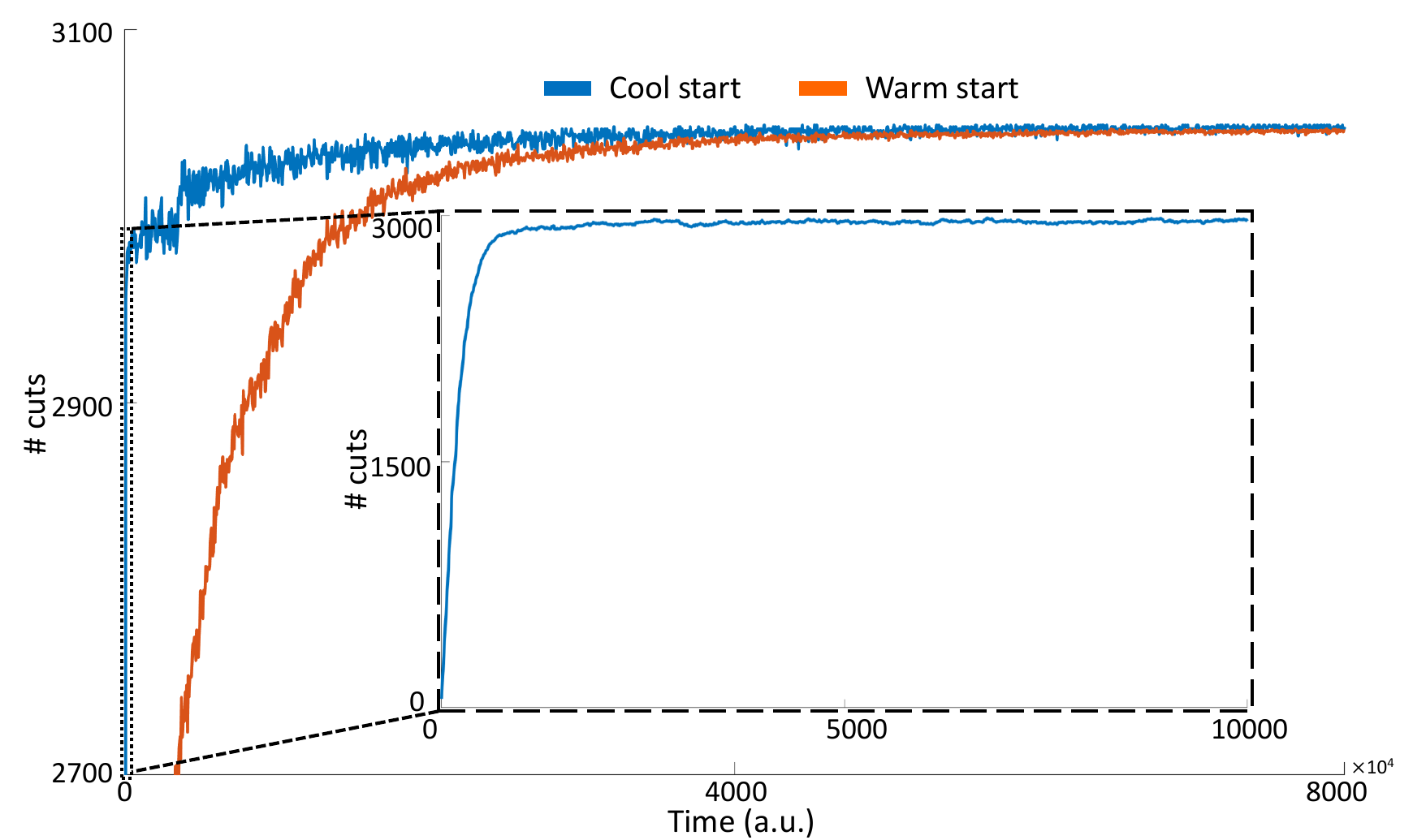}
\caption{Using a low-temperature start to converge fast to a solution neighborhood, after which the annealing is used for exploring the solution space and asymptotic convergence to SOTA solution}\label{fig_supp_coolstart}
\end{figure}
From Eq.~\ref{eq_intro_dH_async} and~\ref{eq_methods_SA_ds}, if NeuroSA is configured with a low-temperature $T_n$ close to $0$, the update of neuron states is determined by the network gradient, which results in faster convergence. This is because during the initial stages of the dynamics, any stochastic factors slows down the convergence. As shown in Fig.~\ref{fig_supp_coolstart}, the cold-start condition pushes the network to convergence to $\sim3000$ cuts for the $G15$ benchmark for which the SOTA is $3050$. However, the dynamics stalls after $\sim10^4$ iterations because the network is trapped in the neighborhood of a local attractor state. However, by re-heating (or adding noise to the threshold) the solution can be further improved, as shown in Fig.~\ref{fig_supp_coolstart}. As indicated in Fig.3e, the time to obtain a unit gain in solution takes up most of the entire duration, at the end of convergence. Therefore, the cold-start strategy accelerates only the initial phase of the optimization, and is intended for practical implementation when there is a simulation time constraint. 
\subsection{Graph Maximum Fanouts/Latency}\label{sec_supp_graph_latency}
\begin{figure}[h]
\centering
\includegraphics[width=0.95\textwidth]{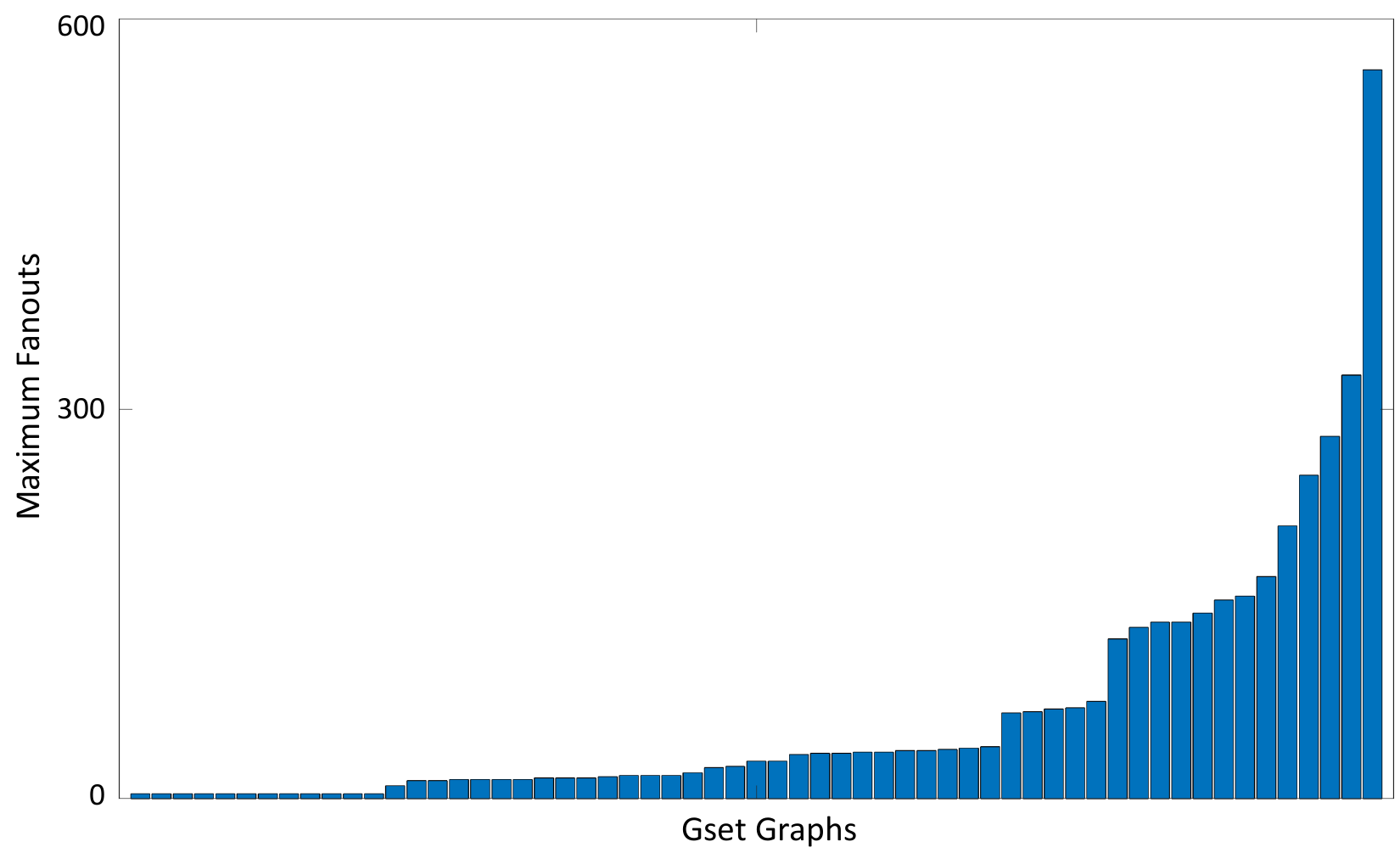}
\caption{Maximum fanouts on each of the Gset benchmarks}\label{fig_supp_graph_latency}
\end{figure}
The latency for propagating spiking events in a conventional sequential implementation of the NeuroSA architecture is determined by the latency between the spike generation to the time when all target neurons receive the spike and estimate the pseudo-gradient. Because of the simplicity of our neuron model design, the delay is mainly embedded in the sequential transmission of the spiking event, due to the shared bus in the system. Therefore, the maximum latency is determined by the largest fanout in the problem graph. As indicated in Fig.~\ref{fig_supp_graph_latency}, the maximum fanout among all tested Gset graphs is around $600$, indicating a consecutive $600$ incidents of routing the spiking events across the system interconnect. On the other hand, in neuromorphic hardware, where the neurons are implemented in parallel and distributed across the network, the transmission delay is offset by the large fanout of the physical interconnects between neuron pairs. Therefore, the neuromorphic implementation is not limited by fanout of the graph. 
\section{Table of Comparison for Gset Benchmarks}\label{sec_supp_8}
The tables~\ref{tab1} and~\ref{tab2} summarize the SOTA solution reported for different Gset benchmarks~\cite{medium2019BenchmarkingMAXCUT} and the difference from the worst-case solution obtained by NeuroSA.
\begin{table}[h]
\caption{Gset1-30 Results}\label{tab1}%
\begin{tabular}{@{}llll@{}}
\toprule
Gset Benchmarks & SOTA Solution & NeuroSA\\
\midrule
G1    & 11624     & 0  \\
G2    & 11620     & -3\\
G3    & 11622     & 0  \\
G4    & 11646     & -5  \\
G5    & 11631     & 0  \\
G6    & 2178     & 0  \\
G7    & 2006     & 0  \\
G10    & 2000   & -1  \\
G11    & 564   & 0  \\
G12    & 556   & 0  \\
G13    & 582   & 0  \\
G14    & 3064   & -1  \\
G15    & 3050   & -1  \\
G16    & 3052   & 0  \\
G17    & 3047   & -2  \\
G18    & 992   & -4  \\
G19    & 906   & -1  \\
G20    & 941   & 0  \\
G21    & 931   & -3  \\
G22    & 13359   & -1  \\
G23    & 13344   & -3  \\
G24    & 13337   & -2  \\
G25    & 13340   & -7  \\
G26    & 13328   & -4  \\
G27    & 3341   & 0  \\
G28    & 3298   & -2 \\
G29    & 3405   & -14  \\
G30    & 3413   & -1  \\

\botrule
\end{tabular}
\end{table}

\begin{table}[h]
\caption{Gset 31-59, 67, 72 Results}\label{tab2}%
\begin{tabular}{@{}llll@{}}
\toprule
Gset Benchmarks & SOTA Solution & NeuroSA\\
\midrule
G31    & 3310   & -2  \\
G32    & 1410   & -4  \\
G33    & 1382   & -2  \\
G34    & 1384   & -2  \\
G35    & 7687   & -12  \\
G36    & 7680   & -17  \\
G37    & 7691   & -8  \\
G38    & 7688   & -16  \\
G39    & 2408   & -3  \\
G40    & 2400   & -7  \\
G41    & 2405   & -12  \\
G42    & 2481   & -15  \\
G43    & 6660   & 0  \\
G44    & 6650   & 0  \\
G45    & 6654   & 0  \\
G46    & 6649   & -3  \\
G47    & 6657   & -1  \\
G48    & 6000   & 0  \\
G49    & 6000   & 0  \\
G50    & 5880   & 0  \\
G51    & 3848   & -1  \\
G52    & 3851   & -4  \\
G53    & 3850   & 0  \\
G54    & 3852   & -4  \\
G55    & 10299   & -15  \\
G56    & 4017   & -11  \\
G57    & 3494   & -22  \\
G58    & 19293   & -39  \\
G59    & 6086   & -31  \\
G67    & 6950   & -68 \\
G72    & 7006   & -76 \\
\botrule
\end{tabular}
\end{table}

\end{document}